\documentclass[sigconf]{acmart}

\usepackage{tabularx}
\usepackage{array}
\usepackage{multirow}
\usepackage{subcaption}
\usepackage{colortbl}
\usepackage{mdframed}

\newcolumntype{Y}{>{\centering\arraybackslash}X}

\providecommand{\rowcolor}[1]{}
\newmdenv[
  leftline=true, rightline=false, topline=false, bottomline=false,
  linewidth=2pt, linecolor=black!70,
  backgroundcolor=black!5,
  innerleftmargin=10pt, innerrightmargin=10pt,
  innertopmargin=6pt, innerbottommargin=6pt,
  skipabove=6pt, skipbelow=6pt,
  nobreak=true
]{promptbox}

\makeatletter
\global\@ACM@balancefalse
\makeatother

\AtBeginDocument{%
  }

\copyrightyear{2026}
\acmYear{2026}
\setcopyright{cc}
\setcctype{by}

\acmConference[MM '26]
{Proceedings of the 34th ACM International Conference on Multimedia}
{November 10--14, 2026}
{Rio de Janeiro, Brazil}

\acmBooktitle{Proceedings of the 34th ACM International Conference on Multimedia
(MM '26), November 10--14, 2026, Rio de Janeiro, Brazil}

\acmDOI{10.1145/3767308.3835998}
\acmISBN{979-8-4007-2213-4/2026/11}
\begin{document}

\title{Simile Understanding in Text-to-Image Models: An Evaluation Framework}

\author{Luecheng Wang}
\orcid{0009-0009-9583-422X}
\email{wanglc@g.ecc.u-tokyo.ac.jp}
\affiliation{%
  \institution{The University of Tokyo}
  \city{Tokyo}
  \country{Japan}}

\author{Shintaro Ozaki}
\orcid{0009-0004-8127-6136}
\email{ozaki.shintaro.ou6@naist.ac.jp}
\affiliation{%
  \institution{Nara Institute of Science and Technology}
  \city{Nara}
  \country{Japan}}

\author{Hidetaka Kamigaito}
\orcid{0000-0002-5249-5813}
\email{kamigaito.h@is.naist.jp}
\affiliation{%
  \institution{Nara Institute of Science and Technology}
  \city{Nara}
  \country{Japan}}

\author{Katsuhiko Hayashi}
\orcid{0000-0002-3240-4697}
\email{katsuhiko-hayashi@g.ecc.u-tokyo.ac.jp}
\affiliation{%
  \institution{The University of Tokyo}
  \city{Tokyo}
  \country{Japan}}

\author{Jingun Kwon}
\orcid{0009-0000-3455-6582}
\email{jingun.kwon@cnu.ac.kr}
\affiliation{%
  \institution{Chungnam National University}
  \city{Daejeon}
  \country{Republic of Korea}}

\author{Manabu Okumura}
\orcid{0009-0001-7730-1536}
\email{oku@lr.pi.titech.ac.jp}
\affiliation{%
  \institution{Institute of Science Tokyo}
  \city{Tokyo}
  \country{Japan}}

\author{Taro Watanabe}
\orcid{0000-0001-8349-3522}
\email{taro@is.naist.jp}
\affiliation{%
  \institution{Nara Institute of Science and Technology}
  \city{Nara}
  \country{Japan}}

\renewcommand{\shortauthors}{Luecheng Wang et al.}

\begin{abstract}
Similes provide a compact and expressive way to describe visual characteristics in text prompts. Recent text-to-image models (t2i models) can produce visually compelling outputs from simile prompts, yet even frontier models frequently misinterpret the metaphorical vehicle and confuse it with the object. These systematic failures reveal a gap between figurative language and object-level visual grounding in t2i models. To investigate this issue, we propose a scalable evaluation framework for simile understanding. Our framework includes (1) a controlled simile dataset in which metaphorical vehicles are drawn from a predefined set of object-detectable categories and combined with diverse templates, (2) automatic grounding metrics based on YOLO (You Only Look Once) detection, and (3) text encoder layer analysis using Diffusion Lens to track how metaphorical vehicles emerge during generation. Experiments across architecturally diverse t2i models reveal consistent literalization failure patterns. We further discuss potential mitigation strategies for improving simile grounding in t2i models.
\end{abstract}



\maketitle

\section{Introduction}

\begin{figure}[t]
\centering
\includegraphics[width=\linewidth]{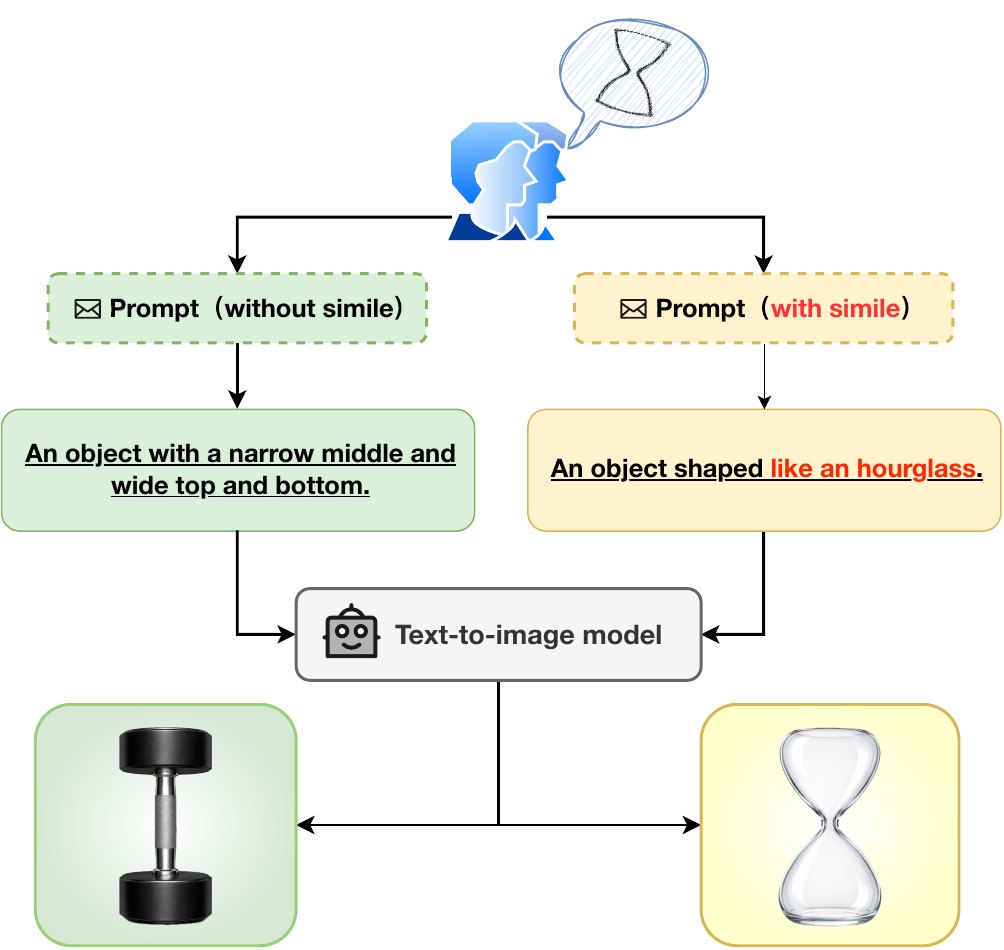}
\caption{
Simile prompts can convey multiple visual properties more concisely than equivalent literal prompts.
}
\label{fig:intro_simile_example}
\Description{
Side-by-side comparison of a literal prompt and a simile prompt describing the same visual properties. The literal prompt lists the properties individually, whereas the simile prompt conveys them compactly through a metaphorical vehicle.
}
\end{figure}

Text-to-image models (t2i models) aim to generate high-quality images from textual prompts~\cite{rombach2022highresolutionimagesynthesislatent,esser2024scalingrectifiedflowtransformers}. As illustrated in Figure~\ref{fig:intro_simile_example}, similes can express visual properties more compactly than literal descriptions. By enabling concise attribute transfer~\cite{lakoff1980metaphors,GENTNER1983155}, similes may therefore serve as a practical tool for prompt design. For example, the simile ``a man is eating bread as hard as stone'' conveys the bread's hardness, rigidity, and resistance without requiring these literal attributes to be listed individually.

Despite this potential, whether t2i models can interpret similes as intended remains underexplored~\cite{kleinlein2022languagedoesdescribelack,yosef-etal-2023-irfl}. It is therefore important to examine the extent to which these models can treat the \emph{metaphorical vehicle}\footnote{In figurative language, the \emph{target} is the entity being described, whereas the \emph{vehicle} is the source whose attributes are transferred to the target.} as a source of attributes rather than as an object to be depicted literally. In the bread example above, the intended property is hardness; an actual stone should not appear in the generated image. We define such literal appearances of the metaphorical vehicle as \textbf{literalization bias}. Evaluating simile understanding in t2i models therefore requires an automatic method for measuring the extent of this bias.

\begin{figure*}[t]
\centering
\includegraphics[width=\textwidth]{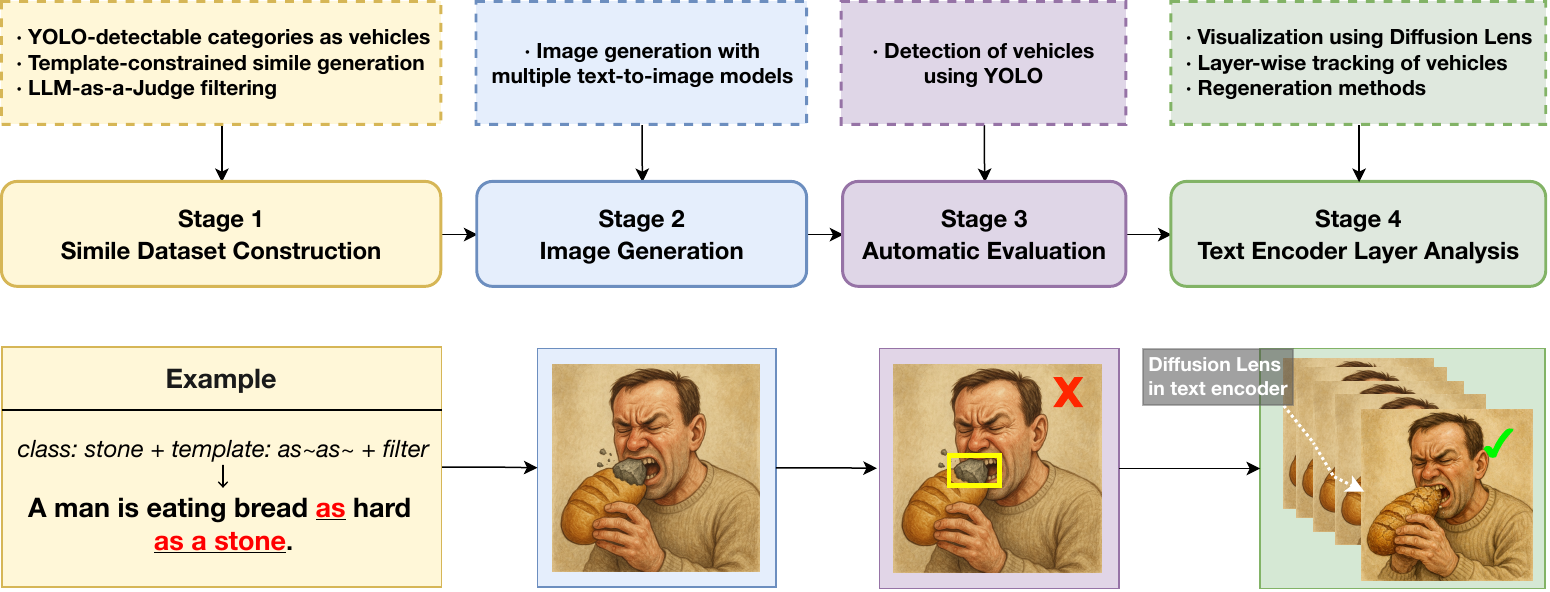}
\caption{
Overview of the proposed framework for evaluating simile understanding in text-to-image models.
}
\label{fig:overview}
\Description{
Overview of the four-stage framework: simile dataset construction, image generation with multiple text-to-image models, automatic detection of literal appearances of metaphorical vehicles, and text encoder layer analysis.
}
\end{figure*}

To this end, we propose an evaluation framework for systematically investigating simile understanding in t2i models (Figure~\ref{fig:overview}). The framework comprises four stages. First, we construct a simile dataset by selecting metaphorical vehicles from a predefined set of object-detectable categories~\cite{rcland12_yolo_classes_2023} and combining them with diverse simile templates. Second, we use the resulting prompts to generate images with multiple t2i models. Third, we apply object detection~\cite{yolo11_ultralytics} to automatically determine whether the metaphorical vehicle appears literally in each generated image~\cite{ghosh2023genevalobjectfocusedframeworkevaluating}. Fourth, we conduct text encoder layer analysis~\cite{toker-etal-2024-diffusion} to track how the metaphorical vehicle emerges throughout the generation process.

Using this evaluation framework, we compare five t2i models with different text encoder configurations. Our results reveal literalization bias across multiple models, indicating a recurring tendency to generate the metaphorical vehicle itself. We also investigate whether this bias can be reduced through two approaches: random regeneration and layer-based regeneration using text encoder layer replacement. Both approaches reduce literalization bias in some models, although their effectiveness varies across models and text encoder configurations.

In summary, the contributions of this paper are threefold:
\begin{itemize}
	\item We define the literal appearance of the metaphorical vehicle in images generated from simile prompts as \textbf{literalization bias} and analyze this tendency across multiple t2i models.
	\item We propose an evaluation framework integrating simile dataset construction using object-detectable categories, image generation with multiple models, automatic evaluation using object detection, and text encoder layer analysis.
	\item We demonstrate that two regeneration methods can partially reduce literalization bias and that their effectiveness varies across models and text encoder configurations.
\end{itemize}

\section{Our Evaluation Framework}

\subsection{Overview}

As shown in Figure~\ref{fig:overview}, the proposed framework comprises four stages, which are summarized below.

\paragraph{Stage 1: Simile Dataset Construction}
We construct a controlled set of simile prompts by selecting metaphorical vehicles from object categories detectable by YOLO (You Only Look Once)~\cite{yolo11_ultralytics} and combining them with predefined simile templates. Multiple large language models (LLMs) generate candidate simile sentences, which are then filtered using multiple LLM-as-a-Judge models~\cite{zheng2023judging,gu2025surveyllmasajudge}.

\paragraph{Stage 2: Image Generation}
Using the simile dataset constructed in Stage 1, we generate images with multiple t2i models~\cite{dreamlike_photoreal_2,flux2024,chen2023pixartalphafasttrainingdiffusion,wu2025qwenimagetechnicalreport,esser2024scalingrectifiedflowtransformers} under a common set of inference settings.

\paragraph{Stage 3: Automatic Evaluation}
We automatically evaluate whether the metaphorical vehicle appears as a literal object in each generated image. For the images produced by each model, we use YOLO~\cite{yolo11_ultralytics} to detect the presence or absence of the metaphorical vehicle and calculate the metaphorical vehicle detection rate.

\paragraph{Stage 4: Text Encoder Layer Analysis}
Using Diffusion Lens~\cite{toker-etal-2024-diffusion}, we obtain layer-wise visualizations from the representations at each text encoder layer and analyze how metaphorical vehicles emerge across layers. We further evaluate random regeneration and layer-based regeneration as potential strategies for reducing literalization bias.

\subsection{Simile Dataset Construction}
\label{sec:template_design}

\subsubsection{Metaphorical Vehicle Selection}

To enable the automatic detection of literal appearances of metaphorical vehicles, we restrict the metaphorical vehicle vocabulary to nouns corresponding to object categories~\cite{rcland12_yolo_classes_2023} detectable by the pretrained YOLO object detector~\cite{yolo11_ultralytics}. Specifically, each metaphorical vehicle is selected from a predefined vocabulary aligned with YOLO detection labels. This design enables us to identify and quantify literalization bias by detecting whether the metaphorical vehicle appears as a literal object in the generated image. We use 80 YOLO-detectable object categories as candidate metaphorical vehicles (Appendices~\ref{sec:app_dataset}) .

\subsubsection{Simile Template Design}

To avoid dependence on any single simile construction, we use multiple templates to generate candidate simile sentences. Specifically, we design 14 simile templates covering comparative constructions, such as \textit{like} and \textit{as \ldots as}, and hypothetical constructions, such as \textit{as if} and \textit{as though}. We also include variants such as \textit{just like}, \textit{exactly like}, and \textit{look like} to increase surface-form diversity.

To maintain grammaticality, metaphorical vehicle nouns are realized in either singular or plural form as required by each template. For nouns that require idiomatic forms, such as \textit{a pair of scissors}, the appropriate forms are specified individually. We further constrain the templates so that the metaphorical vehicle does not serve as the subject of the sentence but instead occurs after a simile marker such as \textit{like} or \textit{as if}. These constraints are designed to produce candidate sentences in which the metaphorical vehicle functions as the source of attribute transfer while preserving variation across simile constructions.

\subsubsection{Simile Generation and Filtering}

Candidate simile sentences are generated using four LLMs: Llama-3.1-8B-Instruct~\cite{grattafiori2024llama3herdmodels}, Gemma-2-9B-IT~\cite{gemmateam2024gemma2improvingopen}, Mistral-7B-Instruct-v0.3~\cite{jiang2023mistral7b}, and Phi-4-mini-instruct~\cite{abdin2024phi4technicalreport}. We generate multiple candidate simile sentences for each combination of metaphorical vehicle and simile template using different random seeds, remove duplicates, verify template conformity, and normalize surface forms.

Each candidate sentence is then evaluated by the three LLMs other than the model that generated it. The judges assess four binary criteria: grammaticality, the presence of a simile, the appropriate use of the metaphorical vehicle, and scene concreteness (Appendices~\ref{sec:app_prompts}). Each positive judgment contributes one point, yielding a maximum agreement score of 12. Sentences receiving the maximum score form the maximum-agreement group (G5). Generation and filtering continue until G5 covers all combinations of 80 metaphorical vehicle categories and 14 templates.

\subsection{Image Generation}

Using the constructed simile dataset, we generate images with five t2i models: Dreamlike-photoreal-2.0~\cite{dreamlike_photoreal_2} (Dreamlike), PixArt-XL-2-1024-MS~\cite{chen2023pixartalphafasttrainingdiffusion} (PixArt), FLUX.1-dev~\cite{flux2024} (FLUX), Stable Diffusion 3.5 Large~\cite{esser2024scalingrectifiedflowtransformers} (SD3.5), and Qwen-Image~\cite{wu2025qwenimagetechnicalreport}. These models cover different text encoder configurations: Dreamlike is CLIP-based~\cite{radford2021learningtransferablevisualmodels}, PixArt is T5-based~\cite{raffel2020exploring}, FLUX and SD3.5 use both CLIP and T5 text encoders, and Qwen-Image uses an LLM-based text encoder.

We provide the same simile prompts to all models and keep the inference settings as consistent as possible. Images are generated at a resolution of $768\times768$ using seed 42, a guidance scale of 4.5, and 30 inference steps. All other model-specific settings follow the default configurations provided by Diffusers.

\subsection{Evaluation}

\subsubsection{Automatic Evaluation}

We measure literalization bias using YOLO-based metaphorical vehicle detection~\cite{yolo11_ultralytics} with the default confidence threshold of 0.25. For each model, we compute \textbf{YOLO-Det}, the proportion of generated images in which YOLO detects the object category corresponding to the metaphorical vehicle. For images with a detected metaphorical vehicle, we additionally record two auxiliary metrics: \textbf{YOLO-Conf}, the confidence score of the corresponding detection, and \textbf{YOLO-Area}, the proportion of the image area covered by its bounding box.

\subsubsection{Human Evaluation}

To assess the validity of the automatic evaluation, we conduct a human evaluation of the generated prompt--image pairs. Annotators score three dimensions on five-point scales: Q1 (Context) for sentence-level metaphor appropriateness, Q2 (Simile) for metaphorical vehicle presence, and Q3 (Relevance) for overall prompt--image semantic alignment.

\subsubsection{Comparison of Evaluation Metrics}

To assess the suitability of different metrics for evaluating literalization bias, we compare YOLO-based metaphorical vehicle detection with two existing image evaluation metrics, CLIPScore and PickScore. We analyze the correlations between the automatic metrics and the human ratings to determine how well each metric captures the literal appearance of the metaphorical vehicle and the overall correspondence between the simile sentence and the generated image.

\subsection{Text Encoder Layer Analysis and Regeneration Methods}
\label{sec:encoder_layer_analysis}

Automatic evaluation of the final generated images does not reveal how metaphorical vehicle appearance varies across text encoder layers. We therefore use Diffusion Lens~\cite{toker-etal-2024-diffusion} to obtain layer-wise visualizations and analyze metaphorical vehicle appearance at different layers. We also evaluate two regeneration methods designed to reduce literalization bias.

\subsubsection{Layer-Wise Analysis}

Using Diffusion Lens~\cite{toker-etal-2024-diffusion}, we obtain layer-wise visualizations from the representations produced at different text encoder layers. These visualizations allow us to examine the layers at which the metaphorical vehicle is depicted as a literal object and to analyze how literalization bias changes across layers.

\subsubsection{Regeneration Methods}
\label{sec:regen_strategies}

To investigate whether literalization bias can be reduced, we evaluate two methods: \textbf{random regeneration} and \textbf{layer-based regeneration}. Neither method modifies the original prompt. For both methods, we apply the same YOLO-based criterion to the initial and regenerated images to determine whether the metaphorical vehicle is detected.

In random regeneration, we repeatedly generate images from the same simile prompt while varying only the random seed. This method serves as a baseline for determining whether literalization bias can be reduced through variation in sampling alone. For each simile prompt, we perform up to five regeneration attempts. After each attempt, YOLO determines whether the metaphorical vehicle is present. The first image in which the metaphorical vehicle is not detected is retained as the regenerated output. If the metaphorical vehicle is detected in all five attempts, the output from the final attempt is retained.

In layer-based regeneration, we vary the text encoder layer used to condition image generation rather than the random seed. For each simile prompt, we first generate an image using the representation from the final text encoder layer. If YOLO detects the metaphorical vehicle, we perform up to five regeneration attempts using representations from successively shallower layers. The first image in which the metaphorical vehicle is not detected is retained as the regenerated output. If no such image is obtained, we retain the original image generated from the final-layer representation.

\section{Evaluation}
\label{sec:evaluation}

\subsection{Evaluation of Simile Dataset}

The simile construction pipeline produced 9,108 judge-evaluated candidate sentences, of which 1,576 received the maximum agreement score and were included in G5. Within G5, the frequencies of the 14 templates range from 85 to 136. The frequencies of the 80 metaphorical vehicle categories range from 14 to 36, with a mean of 19.7 and a standard deviation of 3.89. These distributions indicate that the subset is not dominated by a small number of templates or metaphorical vehicle categories (Appendices~\ref{sec:app_distribution}) .

To examine whether LLM agreement reflects sentence quality, we divide the candidate simile sentences into five agreement groups: G1 (scores 3--6), G2 (7--8), G3 (9--10), G4 (11), and G5 (12). We conduct human evaluation on a stratified sample of 100 sentences covering all 80 metaphorical vehicle categories and all 14 templates. Three annotators rate the same four criteria used by the LLM-as-a-Judge models~\cite{zheng2023judging,gu2025surveyllmasajudge} on a five-point scale.

G5 achieves the highest mean human score among the five groups (Appendices~\ref{sec:app_data_evaluation}) . Although the omnibus ANOVA does not reach significance ($p=0.11$), G5 receives significantly higher ratings than the remaining groups in a focused comparison (Welch's $t=2.09$, $p<0.05$; Mann--Whitney $p<0.02$; Cohen's $d=0.46$). We therefore use maximum agreement as a filtering criterion and restrict image generation experiments to G5.

\subsection{Evaluation of Text-to-Image Models}

\subsubsection{Automatic Evaluation}

We evaluate five t2i models with different text encoder configurations. Table~\ref{tab:model_scores} presents the automatic evaluation results. Lower YOLO-Det values indicate less literalization bias. The detection rates vary substantially across models. Qwen-Image has the highest rate at 0.614, followed by PixArt at 0.495. FLUX, SD3.5, and Dreamlike exhibit lower rates of 0.349, 0.325, and 0.298, respectively. 

Metaphorical vehicles are detected across all 14 template types and in 78 of the 80 metaphorical vehicle categories. Moreover, the 10 most frequently detected categories account for only 22.4\% of all detections, indicating that the observed literalization bias is not concentrated in a small number of templates or metaphorical vehicle categories.

By contrast, CLIPScore and PickScore show relatively limited variation across models. These results suggest that general-purpose image evaluation metrics alone may not adequately capture the literal appearance of metaphorical vehicles.

\begin{table}[t]
\caption{
Automatic evaluation results for five t2i models. Lower YOLO-Det values indicate less literalization bias.
}
\label{tab:model_scores}
\centering
\small
\setlength{\tabcolsep}{3pt}
\renewcommand{\arraystretch}{1.15}
\rowcolors{2}{white}{gray!10}
\begin{tabularx}{\columnwidth}{lYYY}
\toprule
\textbf{Model} & \textbf{YOLO-Det} ($\downarrow$) & \textbf{CLIPScore} ($\uparrow$) & \textbf{PickScore} ($\uparrow$) \\
\midrule
Dreamlike  & \textbf{0.298} & 30.12 & 20.78 \\
FLUX       & 0.349 & 29.03 & 21.26 \\
PixArt     & 0.495 & \textbf{30.59} & \textbf{21.62} \\
Qwen-Image & 0.614 & 30.17 & 21.31 \\
SD3.5      & 0.325 & 29.45 & 20.65 \\
\bottomrule
\end{tabularx}
\end{table}

\begin{table}[t]
\caption{
Human evaluation results for Q2 (Simile) and Q3 (Relevance). Values are means with 95\% bootstrap CIs over 100 images per model, each rated by three annotators.
}
\label{tab:human_q2_q3_scores}
\centering
\small
\setlength{\tabcolsep}{8pt}
\renewcommand{\arraystretch}{1.15}
\rowcolors{2}{white}{gray!10}
\begin{tabularx}{\columnwidth}{lYY}
\toprule
\textbf{Model} & \textbf{Q2 (Simile)} ($\downarrow$) & \textbf{Q3 (Relevance)} ($\uparrow$) \\
\midrule
Dreamlike  & \textbf{2.27} [2.09, 2.46] & 3.45 [3.33, 3.58] \\
FLUX       & 2.28 [2.08, 2.48] & \textbf{3.48} [3.35, 3.61] \\
PixArt     & 3.35 [3.15, 3.56] & 3.21 [3.08, 3.34] \\
Qwen-Image & 3.96 [3.78, 4.14] & 2.85 [2.72, 2.98] \\
SD3.5      & 2.51 [2.31, 2.72] & 3.16 [3.02, 3.30] \\
\bottomrule
\end{tabularx}
\end{table}

\subsubsection{Human Evaluation}

To assess whether the automatic evaluation aligns with human judgments, we conduct a human evaluation of generated prompt--image pairs. We sample 100 sentences from the G5 subset such that all 80 metaphorical vehicle categories and all 14 simile templates are represented. For each sentence, we evaluate the images generated by the five t2i models, yielding a total of 500 prompt--image pairs.

Each pair was evaluated by the same three graduate student annotators recruited from the university, who provided informed consent and were compensated for their participation. One annotator was a native speaker of English, and the other two were English users at the CEFR C1 level. All three annotators had received training in linguistics or a related field. They rated the following items on a five-point scale:

\begin{itemize}
    \item \textbf{Q1 (Context):} How appropriate is the metaphor in the sentence itself?
    \item \textbf{Q2 (Simile):} Is the metaphorical vehicle clearly present in the generated image?
    \item \textbf{Q3 (Relevance):} How well does the generated image reflect the sentence meaning as a whole?
\end{itemize}

The identities of the models were concealed from the annotators, and the generated images were labeled A--E. The mapping between these labels and the models was varied for each sentence, and the presentation order was randomized. Annotators were instructed that there were no objectively correct or incorrect answers and that their ratings should be based on the meaning of the sentence and the presence of the metaphorical vehicle rather than on overall image quality or personal preference. They were not informed of the hypotheses of this study (Appendices~\ref{sec:app_instruction}).

Table~\ref{tab:human_q2_q3_scores} presents the mean scores for Q2 (Simile) and Q3 (Relevance) by model. Q2 measures the perceived presence of the metaphorical vehicle in the generated image, with lower scores indicating that the metaphorical vehicle is less clearly visible. It is therefore the human evaluation item most directly aligned with YOLO-based metaphorical vehicle detection~\cite{yolo11_ultralytics}.
Inter-annotator agreement for Q2 is high, indicating stable judgments of metaphorical vehicle presence (mean pairwise Spearman correlation $\rho = 0.905$; ICC(2,k) $= 0.971$).

Qwen-Image obtains the highest Q2 score at 3.96, followed by PixArt at 3.35. Dreamlike, FLUX, and SD3.5 receive lower scores of 2.27, 2.28, and 2.51, respectively. Q3 measures how well the generated image reflects the overall meaning of the simile sentence, with higher scores indicating stronger correspondence. FLUX and Dreamlike obtain the highest Q3 scores, whereas Qwen-Image receives the lowest score at 2.85. Across the five models, higher Q2 scores tend to coincide with lower Q3 scores.

\subsection{Comparison of Evaluation Metrics}

To assess how well different metrics capture literalization bias, we compare the YOLO-based metrics with two general-purpose image evaluation metrics, CLIPScore and PickScore, using human ratings as references.

Table~\ref{tab:metric_alignment} reports image-level correlations, measured using Pearson's $r$ and Spearman's $\rho$, between each automatic metric and the human ratings across 500 prompt--image pairs. YOLO-Det shows a strong correlation with Q2 (Simile) ($r=0.826$, $\rho=0.798$), while YOLO-Area also correlates positively with Q2. Among images with a detected metaphorical vehicle, mean YOLO-Area ranges from 0.116 to 0.147 and mean YOLO-Conf from 0.737 to 0.819 across models (Appendices~\ref{sec:app_intensity_stats}) . Thus, the main cross-model variation lies in how frequently metaphorical vehicles are detected rather than in their detected area or confidence.

\begin{figure}[t]
\centering
\begin{subfigure}[t]{0.49\linewidth}
\centering
\includegraphics[width=\linewidth]{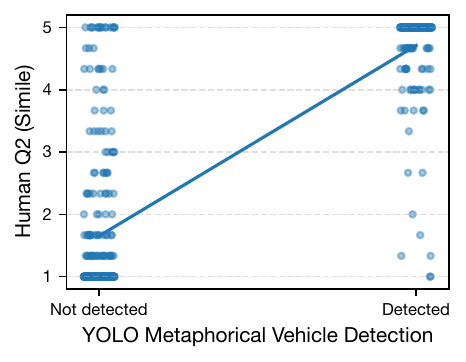}
\caption{Image level}
\label{fig:auto_human_image}
\end{subfigure}
\hfill
\begin{subfigure}[t]{0.49\linewidth}
\centering
\includegraphics[width=\linewidth]{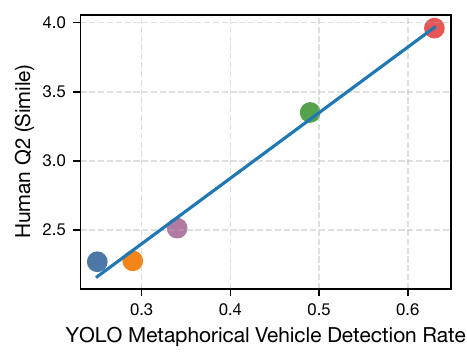}
\caption{Model level}
\label{fig:auto_human_model}
\end{subfigure}
\caption{
Agreement between YOLO-based metaphorical vehicle detection and human Q2 (Simile) ratings at the image and model levels.
(a) Mean Q2 ratings for images with and without YOLO detection.
(b) Relationship between YOLO-Det and mean Q2 ratings across the five t2i models.
}
\Description{
Panel (a) compares Q2 ratings for images with and without YOLO detections of the metaphorical vehicle. Panel (b) compares each model's YOLO-Det rate with its mean Q2 rating.
}
\end{figure}

As shown in Figure~\ref{fig:auto_human_image}, images in which YOLO detects the metaphorical vehicle receive substantially higher Q2 (Simile) scores than images without such detections, with mean scores of 4.71 and 1.65, respectively (Welch's $t=35.9$, $p<0.001$, Cohen's $d=2.99$). This result indicates strong agreement between YOLO-based detection and human judgments of metaphorical vehicle presence. At the model level, as shown in Figure~\ref{fig:auto_human_model}, the model ranking based on YOLO-Det is nearly identical to that based on the mean Q2 score ($r=0.99$, $\rho=1.00$).

By contrast, CLIPScore and PickScore show substantially weaker correlations with Q2 than YOLO-Det, and their correlations with Q3 (Relevance) are close to zero. The YOLO-based metrics show moderate correlations with Q3. These results indicate that general-purpose image evaluation metrics do not adequately capture either the literal appearance of metaphorical vehicles or the overall correspondence between simile prompts and generated images.

Human ratings for Q2 (Simile) and Q3 (Relevance) are strongly negatively correlated ($r=-0.601$, $\rho=-0.607$). Thus, images in which the metaphorical vehicle is rated as more clearly visible tend to receive lower ratings for overall correspondence with the simile sentence.

\begin{table}[t]
\caption{
Image-level correlations between automatic metrics and human ratings. For Q3 (Relevance), the signs of the YOLO-based correlations are reversed so that higher values indicate stronger agreement.
}
\label{tab:metric_alignment}
\centering
\small
\setlength{\tabcolsep}{5pt}
\renewcommand{\arraystretch}{1.15}
\rowcolors{2}{white}{gray!10}
\begin{tabular}{lcc}
\toprule
\textbf{Evaluation Metric} & \textbf{Q2 (Simile)} ($\uparrow$) & \textbf{Q3 (Relevance)} ($\uparrow$) \\
\midrule
YOLO-Det   & $r=\textbf{0.826},\ \rho=\textbf{0.798}$ & $r=\textbf{0.509},\ \rho=0.517$ \\
YOLO-Area  & $r=0.437,\ \rho=0.772$ & $r=0.423,\ \rho=\textbf{0.545}$ \\
CLIPScore  & $r=0.382,\ \rho=0.394$ & $r=-0.043,\ \rho=-0.095$ \\
PickScore  & $r=0.342,\ \rho=0.353$ & $r=-0.032,\ \rho=-0.064$ \\
Human (Q2 vs.\ Q3) & \multicolumn{2}{c}{$r=-0.601,\ \rho=-0.607$} \\
\bottomrule
\end{tabular}
\end{table}

\subsection{Regeneration Results}
\label{sec:regen_eval}

We evaluate whether random regeneration and layer-based regeneration can reduce literalization bias without modifying the original prompt. Table~\ref{tab:regeneration_results} summarizes the results of both methods.

\begin{table}[t]
\caption{
Results of the two regeneration methods. $\Delta$ YOLO-Det denotes the change relative to initial generation, with more negative values indicating larger reductions.
}
\label{tab:regeneration_results}
\centering
\footnotesize
\setlength{\tabcolsep}{2.5pt}
\renewcommand{\arraystretch}{1.15}

\definecolor{groupgray}{gray}{0.94}
\definecolor{groupblue}{HTML}{EEF5FF}

\begin{tabular}{lcccc}
\toprule
\textbf{Model} & \textbf{YOLO-Det} ($\downarrow$)& \textbf{$\Delta$ YOLO-Det} ($\downarrow$) & \textbf{CLIPScore} ($\uparrow$) & \textbf{PickScore} ($\uparrow$)  \\
\midrule
\rowcolor{groupgray} \multicolumn{5}{l}{\textbf{Random Regeneration}} \\
Dreamlike    & \textbf{0.124} & -0.174 & 29.81 & 20.71 \\
PixArt       & 0.308 & -0.187 & \textbf{30.45} & \textbf{21.52} \\
Qwen-Image   & 0.244 & \textbf{-0.370} & 29.36 & 20.99 \\
FLUX         & 0.178 & -0.171 & 28.74 & 21.18 \\
SD3.5        & 0.168 & -0.157 & 29.27 & 20.60 \\

\addlinespace[2pt]
\rowcolor{groupblue} \multicolumn{5}{l}{\textbf{Layer-Based Regeneration}} \\
Dreamlike      & \textbf{0.089} & -0.209 & 29.51 & 20.63 \\
PixArt         & 0.176 & -0.319 & \textbf{30.28} & \textbf{21.54} \\
Qwen-Image     & 0.255 & \textbf{-0.359} & 28.95 & 20.87 \\
FLUX (T5)      & 0.235 & -0.114 & 28.87 & 21.22 \\
FLUX (CLIP)    & 0.271 & -0.078 & 28.89 & 21.22 \\
SD3.5 (T5)     & 0.214 & -0.111 & 29.33 & 20.55 \\
SD3.5 (CLIP-L) & 0.222 & -0.103 & 29.24 & 20.53 \\
SD3.5 (CLIP-G) & 0.225 & -0.100 & 29.23 & 20.53 \\
\bottomrule
\end{tabular}
\end{table}

\subsubsection{Random Regeneration}

Random regeneration reduces YOLO-Det for all five models. Relative to initial generation, the reductions range from $-0.157$ for SD3.5 to $-0.370$ for Qwen-Image. These results show that literalization bias can sometimes be reduced through repeated sampling with different random seeds. Qwen-Image and PixArt, which exhibit relatively high YOLO-Det values under initial generation, show particularly large reductions. By contrast, the reductions are smaller for Dreamlike and SD3.5.

\subsubsection{Layer-Based Regeneration}

Layer-based regeneration reduces YOLO-Det relative to initial generation across all evaluated models and text encoder branches. However, the magnitude of the reduction varies substantially across models and text encoder branches. For Dreamlike and PixArt, regeneration using representations from shallower layers produces large reductions, with $\Delta$ YOLO-Det values of $-0.209$ and $-0.319$, respectively. Qwen-Image likewise shows a large reduction of $-0.359$, although its post-regeneration YOLO-Det remains relatively high at 0.255.

The reductions are more limited for FLUX and SD3.5. For FLUX, $\Delta$ YOLO-Det ranges from $-0.078$ to $-0.114$, while for SD3.5 it ranges from $-0.100$ to $-0.111$. Within FLUX, regeneration through the T5 branch yields a lower YOLO-Det value than regeneration through the CLIP branch. For SD3.5, the differences among the T5, CLIP-L, and CLIP-G branches are small.

\subsubsection{Comparison of Regeneration Methods}

Both regeneration methods reduce literalization bias, but their effects differ across models. Random regeneration lowers YOLO-Det for every model. Layer-based regeneration also lowers YOLO-Det under all evaluated conditions, but its effectiveness is strongly model-dependent. These results characterize random regeneration as a broadly applicable baseline, whereas the effectiveness of layer-based regeneration depends on the model and text encoder branch.

CLIPScore and PickScore change only modestly across regeneration conditions, even when YOLO-Det decreases substantially. This finding further indicates that these general-purpose metrics alone are insufficient for measuring reductions in literalization bias.

\section{Text Encoder Layer Analysis}
\label{sec:diffusion}

\begin{figure*}[t]
\centering
\begin{subfigure}[t]{0.32\textwidth}
\centering
\includegraphics[width=\linewidth]{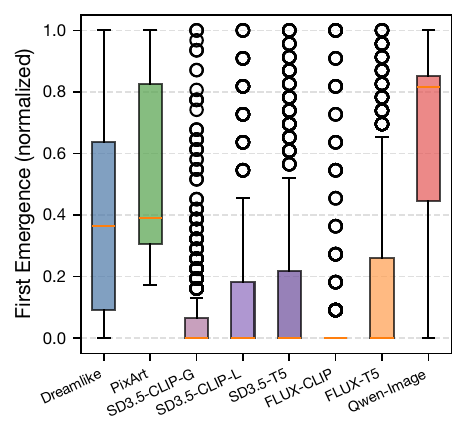}
\caption{First-emergence layer}
\label{fig:trajectory_first_emergence}
\end{subfigure}
\hfill
\begin{subfigure}[t]{0.32\textwidth}
\centering
\includegraphics[width=\linewidth]{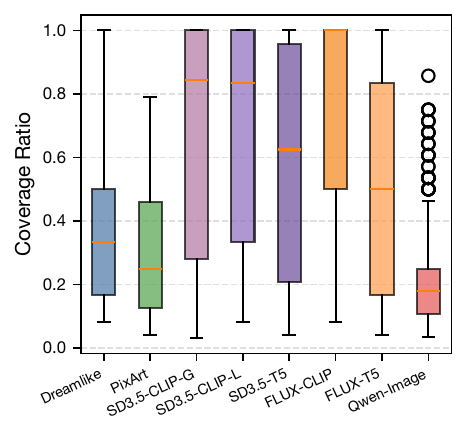}
\caption{Coverage ratio}
\label{fig:trajectory_coverage_ratio}
\end{subfigure}
\hfill
\begin{subfigure}[t]{0.32\textwidth}
\centering
\includegraphics[width=\linewidth]{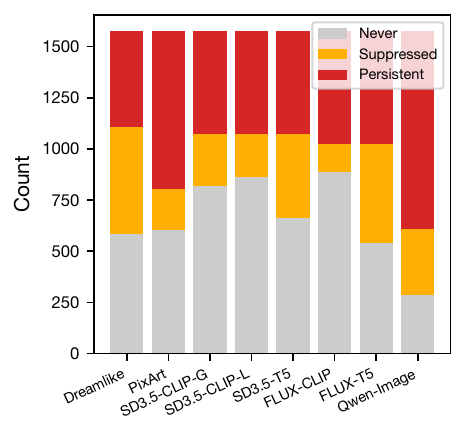}
\caption{Appearance category}
\label{fig:trajectory_process_breakdown}
\end{subfigure}
\caption{
Text encoder layer analysis of metaphorical vehicle appearance.
(a) Normalized first-emergence layer, where lower values indicate detection in shallower layers.
(b) Coverage ratio, defined as the proportion of text encoder layers in which the metaphorical vehicle is detected.
(c) Distribution of the \textit{Never}, \textit{Suppressed}, and \textit{Persistent} appearance categories.
}
\label{fig:trajectory_analysis}
\Description{
Three plots summarizing the literal appearance of metaphorical vehicles across text encoder layers. Panel (a) shows the normalized layer at which each metaphorical vehicle is first detected, with smaller values indicating shallower layers. Panel (b) shows the proportion of text encoder layers in which the metaphorical vehicle is detected. Panel (c) shows three appearance categories: Never, when the metaphorical vehicle is not detected; Suppressed, when it is detected in layer-wise visualizations but not in the final image; and Persistent, when it is detected both in layer-wise visualizations and in the final image.
}
\end{figure*}

To examine how literal depictions of metaphorical vehicles vary across text encoder layers, we apply Diffusion Lens~\cite{toker-etal-2024-diffusion} to obtain layer-wise visualizations. For each model and text encoder branch, we use YOLO~\cite{yolo11_ultralytics} to determine whether the metaphorical vehicle is present in the visualization produced from each layer representation. We then calculate the following three metrics:

\begin{itemize}
    \item \textbf{First-emergence layer:} The earliest layer at which the metaphorical vehicle is detected, normalized to the range $[0,1]$ as $\frac{\text{layer}-1}{\text{total layers}-1}$. 
    \item \textbf{Coverage ratio:} The proportion of text encoder layers in which the metaphorical vehicle is detected.
    \item \textbf{Appearance category:} A three-way classification of whether the metaphorical vehicle (i) never appears (\textit{Never}), (ii) appears in layer-wise visualizations but is absent from the final image (\textit{Suppressed}), or (iii) appears in layer-wise visualizations and is also present in the final image (\textit{Persistent}).
\end{itemize}

This analysis extends the final-image evaluation by showing where literal depictions of metaphorical vehicles first emerge, how broadly they occur across text encoder layers, and whether they remain in the final output. The results are presented in Figure~\ref{fig:trajectory_analysis}.

\subsection{First-Emergence Layer}

Figure~\ref{fig:trajectory_first_emergence} shows the normalized first-emergence layer for each model and text encoder branch. The values vary substantially across models and text encoder branches. The CLIP-based branches of SD3.5 and FLUX exhibit particularly early emergence, with mean normalized values of 0.08 for both SD3.5-CLIP-G and FLUX-CLIP. 
By contrast, Qwen-Image exhibits the latest emergence, with a mean of approximately 0.68, followed by PixArt at approximately 0.54. Dreamlike has an intermediate mean of approximately 0.37, whereas the T5-based branches show earlier emergence, with means ranging from approximately 0.17 to 0.20.

\subsection{Coverage Ratio}

Figure~\ref{fig:trajectory_coverage_ratio} shows the proportion of text encoder layers in which the metaphorical vehicle is detected. The CLIP-based branches exhibit high coverage ratios, with mean values ranging from approximately 0.67 to 0.77, indicating that metaphorical vehicles are detected across a large proportion of layers. 
By contrast, Qwen-Image has the lowest coverage ratio, with a mean of approximately 0.19. Together with its late first-emergence layer, this result indicates that metaphorical vehicle detection in Qwen-Image is concentrated within a limited range of deeper layers. PixArt and Dreamlike show intermediate coverage ratios, with means of approximately 0.31 and 0.35, respectively. The T5-based branches also exhibit intermediate values, lower than those of the CLIP-based branches but higher than that of Qwen-Image.

\subsection{Appearance Tendencies}

Figure~\ref{fig:trajectory_process_breakdown} shows the proportions of the three appearance categories: \textit{Never}, \textit{Suppressed}, \textit{Persistent}. When considered together with the first-emergence layer and coverage ratio, these results reveal three broader tendencies in the literal appearance of metaphorical vehicles: (1) appearance in shallow layers followed by persistence to the final image, (2) appearance in intermediate layers followed by suppression before the final image, and (3) appearance in deeper layers followed by persistence to the final image (Appendices~\ref{sec:app_difflens}) .

First, the CLIP-based branches contain many cases in which metaphorical vehicles appear in shallow layers and persist to the final image. For FLUX-CLIP, the mean first-emergence layer is 0.080 and the mean coverage ratio is 0.768, with 550 Persistent cases and 138 Suppressed cases. Similarly, SD3.5-CLIP-G and SD3.5-CLIP-L have shallow mean first-emergence layers of 0.080 and 0.145, respectively, together with high mean coverage ratios of 0.667 and 0.677. These results indicate that, in the CLIP-based branches, metaphorical vehicles tend to appear early across text encoder layers and persist to the final image.

Second, in Dreamlike, metaphorical vehicles often appear in intermediate layers but not in the final image. Its mean first-emergence layer is 0.373 and its mean coverage ratio is 0.354, with 522 Suppressed cases and 469 Persistent cases. Thus, even when the metaphorical vehicle appears in the layer-wise visualizations, it is frequently absent from the final image. Substantial numbers of Suppressed cases are also observed in the T5-based branches, with 411 cases for SD3.5-T5 and 486 for FLUX-T5.

Third, Qwen-Image contains many cases in which metaphorical vehicles appear in deeper layers and persist to the final image.
Its mean first-emergence layer is relatively late at 0.679, and its mean coverage ratio is low at 0.187, with 967 Persistent cases and 323 Suppressed cases. PixArt exhibits a similar tendency, with a relatively late mean first-emergence layer of 0.542, 770 Persistent cases, and 203 Suppressed cases. Thus, although metaphorical vehicle appearances are concentrated within a limited range of deeper layers, they frequently persist to the final image.

These findings show that literalization bias is related not only to the metaphorical vehicle detection rate in the final image but also to differences in the first-emergence layer, coverage ratio, and appearance tendencies across text encoder configurations.

\subsection{Implications and Limitations}

The results above indicate that the literal appearance of the metaphorical vehicle is closely associated with the overall correspondence between simile sentences and generated images. Human ratings for Q2 (Simile) and Q3 (Relevance) show a strong negative correlation: as the metaphorical vehicle becomes more clearly visible, annotators tend to judge the overall correspondence between the sentence and the image as lower. This pattern is consistent with the model treating the metaphorical vehicle as an object to depict rather than as a source of attributes for the target. Its literal depiction therefore constitutes a salient and measurable failure mode in image generation from simile prompts.

However, the evaluated metrics differ substantially in their ability to capture this failure. YOLO-based metaphorical vehicle detection~\cite{yolo11_ultralytics} aligns strongly with human judgments of metaphorical vehicle presence and provides an effective measure of literalization bias. By contrast, CLIPScore and PickScore show only weak correlations with Q2 (Simile) and Q3 (Relevance) and therefore do not adequately capture either the literal appearance of metaphorical vehicles or the overall correspondence between simile sentences and generated images.

These findings clarify both the utility and the limitations of the proposed evaluation framework. Object detection provides an effective means of measuring literalization bias as a clearly observable failure. However, the absence of a detected metaphorical vehicle does not by itself demonstrate that the model has interpreted the simile as intended, because the attributes associated with the metaphorical vehicle may still be insufficiently reflected in the generated image.

Accordingly, reducing literalization bias should be understood as addressing one aspect of failure in image generation from simile prompts. It reduces the literal appearance of the metaphorical vehicle but does not, on its own, demonstrate complete understanding of the simile.

\section{Related Work}

\subsection{Figurative Language in Natural Language Processing}

Processing figurative expressions such as metaphors and similes is a longstanding challenge in natural language processing. Unlike literal language, figurative language requires models to move beyond surface lexical meanings and interpret relationships between a target and another conceptual domain. Metaphors are commonly described in terms of correspondences between conceptual domains~\cite{lakoff1980metaphors,GENTNER1983155}, while similes make such correspondences explicit through comparison markers such as \textit{like} and \textit{as if}. Although these markers make the comparison structure explicit, resolving the intended relationship between the target and the metaphorical vehicle remains a nontrivial interpretation problem~\cite{he-etal-2022-pre}.

Computational research on figurative language has primarily focused on metaphor detection, interpretation, and generation~\cite{tong-etal-2021-recent}. Early work used rule-based and statistical methods to identify metaphorical usage from lexical choices and contextual information~\cite{birke-sarkar-2006-clustering,shutova-2010-models}. Subsequent studies employed neural models and distributed representations to capture semantic variation associated with metaphorical usage~\cite{hamilton-etal-2016-diachronic,gao2018neuralmetaphordetectioncontext}. Metaphor generation has also been studied as a controlled generation problem, using conceptual mappings, symbolic knowledge, and discriminative decoding to produce metaphorical paraphrases and creative text~\cite{stowe-etal-2021-metaphor,stowe-etal-2021-exploring,chakrabarty-etal-2021-mermaid}. Simile research has examined recognition and component extraction~\cite{liu-etal-2018-neural,wang-etal-2022-getting}, interpretation~\cite{he-etal-2022-pre,chen-etal-2022-probing}, and generation from literal text~\cite{chakrabarty-etal-2020-generating}. More recently, large language models have been evaluated on tasks including metaphor detection, interpretation, explanation, paraphrasing, and generation, as well as across different prompting configurations~\cite{distefano2024,ichien2024largelanguagemodeldisplays,tong-etal-2024-metaphor,sanchez-bayona-agerri-2025-metaphor}.

Most of this research, however, focuses on textual judgments or text generation. Typical evaluation criteria include classification accuracy, explanation quality, and the naturalness of generated sentences~\cite{bowman2021fixbenchmarkingnaturallanguage}. Although such evaluations assess whether models can identify or explain figurative expressions in text, they do not directly examine how figurative meaning is reflected in visual outputs. Competence in processing figurative language at the textual level therefore does not necessarily imply the ability to represent the intended attributes appropriately in a visual scene~\cite{bisk-etal-2020-experience}.

This distinction is particularly important for text-to-image generation. When a simile contains a concrete metaphorical vehicle, a t2i model may depict the metaphorical vehicle as a literal object rather than transfer its attributes to the target. Evaluating how metaphorical vehicles appear in generated images is therefore necessary for identifying this form of failure.

\subsection{Figurative Language in Image Generation}

Recent t2i models have achieved substantial improvements in generating high-quality images from textual prompts~\cite{rombach2022highresolutionimagesynthesislatent}. The use of text encoders such as CLIP~\cite{radford2021learningtransferablevisualmodels} and T5~\cite{raffel2020exploring} has also improved their ability to condition image generation on complex natural language descriptions. Prompt refinement can further improve image quality and text--image alignment~\cite{ozaki2026texttigertextbasedintelligentgeneration}. However, systematic evaluation of how t2i models handle simile prompts remains limited~\cite{yosef-etal-2023-irfl,kleinlein2022languagedoesdescribelack}.

Research on visual metaphor examines how metaphorical meaning is encoded in and inferred from images. Earlier studies proposed typologies of visual rhetoric and procedures for identifying visual metaphors in images~\cite{beyondvisualmetaphor,visualmetaphoridentification}. More recent computational work has introduced tasks and models for visual metaphor classification, localization, understanding, generation, and multimodal detection~\cite{akula2023metacluecomprehensivevisualmetaphors,xu-etal-2024-exploring}. These studies generally treat metaphorical meaning as encoded by an existing or deliberately composed image through visual composition or image--text interaction. Our setting instead begins with a linguistic simile and evaluates whether its metaphorical vehicle is rendered as an unintended literal object.

Existing image generation benchmarks primarily evaluate text--image alignment, object presence, compositional relations, and image quality~\cite{ghosh2023genevalobjectfocusedframeworkevaluating,huang2025t2icompbenchenhancedcomprehensivebenchmark,sun2025t2ireasonbenchbenchmarkingreasoninginformedtexttoimage}.
These evaluations are useful for determining whether objects and relations explicitly described in a prompt are represented in the generated image. Simile prompts, however, introduce a different type of failure: the metaphorical vehicle may be rendered as a literal object rather than serving as a source of attributes for the target. General measures of text--image alignment and image quality may therefore fail to capture the literal appearance of metaphorical vehicles.

Research on multimodal reasoning has examined compositional understanding, spatial relations, visual commonsense, and related capabilities~\cite{sun2025t2ireasonbenchbenchmarkingreasoninginformedtexttoimage,wang2026placebenchmarkingspatialintelligence}. In simile prompts, however, a metaphorical vehicle may correspond to a concrete object category, causing a t2i model to depict the object literally. This failure differs from conventional compositional reasoning errors because the object named in the prompt should not necessarily appear in the generated image. Although recent studies have begun to investigate metaphor understanding and figurative reasoning in multimodal models~\cite{kundu-etal-2025-looking,saakyan-etal-2025-understanding}, systematic evaluations of literal metaphorical vehicle appearances in images generated from simile prompts remain limited.

Related work has also explored creative image generation and figurative prompts~\cite{chakrabarty-etal-2023-spy,efficient-visual}. Such studies often emphasize creativity, aesthetic quality, or overall preference, rather than explicitly evaluating whether the metaphorical vehicle is depicted as a literal object~\cite{su2024efficient,koushik2025mindseyemultifacetedreward}. Furthermore, little attention has been paid to how literal depictions of metaphorical vehicles vary across text encoder layers. Methods such as Diffusion Lens~\cite{toker-etal-2024-diffusion} make it possible to examine how representations from individual text encoder layers are reflected in generated images, but their application to metaphorical vehicle appearance in simile prompts remains limited.

In this study, we define the literal appearance of a metaphorical vehicle in images generated from simile prompts as literalization bias and measure its prevalence across t2i models. By combining automatic evaluation using object detection with text encoder layer analysis, we examine both how frequently metaphorical vehicles are depicted literally and how their appearance varies across text encoder layers.

\section{Conclusion}

We define literalization bias as the literal depiction of the metaphorical vehicle in images generated from simile prompts and evaluate this phenomenon across five t2i models. Our results show that literalization bias occurs across multiple models, although its prevalence varies substantially across models and text encoder configurations.

To support this analysis, we propose an evaluation framework that integrates controlled simile dataset construction, image generation with multiple t2i models, automatic evaluation using object detection, and text encoder layer analysis. YOLO-based metaphorical vehicle detection aligns strongly with human judgments of vehicle presence, whereas CLIPScore and PickScore show limited correspondence with both metaphorical vehicle visibility and prompt--image relevance. Analysis using Diffusion Lens~\cite{toker-etal-2024-diffusion} reveals substantial differences across models in the first-emergence layer, coverage ratio, and appearance tendencies of metaphorical vehicles.

Both random regeneration and layer-based regeneration partially reduce literalization bias without modifying the prompt, but their effectiveness is model-dependent. Reducing literalization bias does not demonstrate that the intended attributes have been successfully reflected in the image; a more comprehensive assessment therefore requires evaluation of attribute transfer. Nevertheless, literalization bias provides a measurable dimension for evaluating one aspect of simile understanding in t2i models.

\bibliographystyle{ACM-Reference-Format}
\bibliography{references}

\appendix

\section{Supplementary Experimental Settings}
\label{sec:app_setup}

\begin{table}[h]
\caption{List of the 80 YOLO-detectable categories.}
\label{tab:app_yolo_classes}
\centering
\scriptsize
\setlength{\tabcolsep}{4pt}
\renewcommand{\arraystretch}{1.2}
\begin{tabularx}{\linewidth}{XXXXX}
\toprule
person & bicycle & car & motorcycle & airplane \\
bus & train & truck & boat & traffic light \\
fire hydrant & stop sign & parking meter & bench & bird \\
cat & dog & horse & sheep & cow \\
elephant & bear & zebra & giraffe & backpack \\
umbrella & handbag & tie & suitcase & frisbee \\
skis & snowboard & sports ball & kite & baseball bat \\
baseball glove & skateboard & surfboard & tennis racket & bottle \\
wine glass & cup & fork & knife & spoon \\
bowl & banana & apple & sandwich & orange \\
broccoli & carrot & hot dog & pizza & donut \\
cake & chair & couch & potted plant & bed \\
dining table & toilet & tv & laptop & mouse \\
remote & keyboard & cell phone & microwave & oven \\
toaster & sink & refrigerator & book & clock \\
vase & scissors & teddy bear & hair drier & toothbrush \\
\bottomrule
\end{tabularx}
\end{table}

\begin{table}[h]
\caption{
Full template distribution and the 14 most and least frequent metaphorical vehicle categories in the maximum-agreement group (G5). [s] and [p] denote singular and plural forms, respectively.
}
\label{tab:all-yes_distribution}
\centering
\footnotesize
\setlength{\tabcolsep}{4pt}
\renewcommand{\arraystretch}{1.3}
\begin{tabular*}{\linewidth}{@{\extracolsep{\fill}}>{\columncolor{blue!05}[0pt][\tabcolsep]}l >{\columncolor{blue!05}}r >{\columncolor{orange!10}}l >{\columncolor{orange!10}}r >{\columncolor{orange!10}}l >{\columncolor{orange!10}}r}
\toprule
\multicolumn{2}{>{\columncolor{blue!05}}c}{\textbf{Template Distribution}} & \multicolumn{2}{>{\columncolor{orange!10}}c}{\textbf{Top 14 Categories}} & \multicolumn{2}{>{\columncolor{orange!10}}c}{\textbf{Bottom 14 Categories}} \\
\midrule
like [s]              & 113 & bowl           & 36 & dog          & 16 \\
like [p]              & 136 & wine glass     & 30 & pizza        & 16 \\
just like [s]         & 121 & oven           & 28 & toaster      & 16 \\
just like [p]         & 123 & bottle         & 28 & microwave    & 16 \\
looks like [s]        & 116 & baseball bat   & 26 & carrot       & 16 \\
look like [p]         & 85  & knife          & 25 & book         & 16 \\
exactly like [s]      & 99  & vase           & 25 & laptop       & 15 \\
exactly like [p]      & 99  & bed            & 25 & hot dog      & 15 \\
as $\sim$ as [s]      & 98  & potted plant   & 24 & motorcycle   & 15 \\
as $\sim$ as [p]      & 96  & baseball glove & 24 & fire hydrant & 15 \\
as if $\sim$ [s]     & 132 & bird           & 23 & broccoli     & 15 \\
as if $\sim$ [p]     & 122 & orange         & 23 & cat          & 15 \\
as though $\sim$ [s] & 131 & spoon          & 23 & clock        & 14 \\
as though $\sim$ [p] & 105 & person         & 23 & cow          & 14 \\
\bottomrule
\end{tabular*}
\end{table}

\subsection{Simile Dataset Construction}
\label{sec:app_dataset}

Metaphorical vehicles are selected from a predefined vocabulary aligned with categories detectable by YOLO~\cite{yolo11_ultralytics}. Table~\ref{tab:app_yolo_classes} lists the 80 detectable categories~\cite{rcland12_yolo_classes_2023} used as metaphorical vehicle categories. 

\subsection{Prompts and Setting}
\label{sec:app_prompts}

\paragraph{Candidate simile sentence generation.}
For candidate simile sentence generation, we use a maximum length of 40 tokens, temperature 0.4, top-$p$ 0.9, and top-$k$ 50. Candidate generation is repeated with random seeds varied sequentially from 1. The following fixed prompt is used to generate candidate simile sentences. In each run, the prompt placeholder is replaced with the target template (e.g., \textit{like a car}), and one sentence is generated.

\begin{promptbox}
\footnotesize\ttfamily
Please write one grammatically correct, natural English sentence that uses a simile (limited to 20 words). \\
The sentence must use the following exact template as the simile's vehicle to describe a different main subject, not the vehicle itself, its part, or a similar entity. \\
The sentence must not be a simple statement, but should instead use the simile to enhance the description of the scene. \\
The sentence should describe a concrete, visual scene. \\
Avoid pronouns, emotional states, and broad abstract concepts. \\
Template: ``\{template\_phrase\}'' \\
A:
\end{promptbox}

\paragraph{LLM-as-a-Judge filtering.}
For LLM-as-a-Judge filtering, we use temperature 0 and greedy decoding without sampling. The following fixed prompt is used for LLM-as-a-Judge filtering of candidate simile sentences. In each run, \texttt{\{question\}} is replaced with the question corresponding to an evaluation criterion, and \texttt{\{sentence\}} is replaced with the candidate simile sentence being evaluated.

\begin{promptbox}
\footnotesize\ttfamily
Q: \{question\} \\
Answer only ``yes'' or ``no''. \\
Sentence: ``\{sentence\}'' \\
A:
\end{promptbox}

The following four questions are used for \texttt{\{question\}}. For the question, \texttt{\{vehicle\}} is replaced with the vehicle noun specified in the candidate sentence.
\begin{itemize}
    \item Is the following sentence complete, grammatically correct, and natural?
    \item Is there a simile in the following sentence?
    \item In the following sentence, is ``\{vehicle\}'' only used as the metaphorical vehicle and not the similar entity of the subject?
    \item Does the following sentence describe a concrete, visualizable scene?
\end{itemize}

\section{Supplementary Results}
\label{sec:app_dataset_stats}

\subsection{Distribution of the G5 Subset}
\label{sec:app_distribution}

Table~\ref{tab:all-yes_distribution} provides the full template distribution and the 14 most and least frequent metaphorical vehicle categories in the maximum-agreement group (G5).

\subsection{Simile Dataset Evaluation}
\label{sec:app_data_evaluation}

\begin{figure}[t]
\centering
\begin{subfigure}[t]{0.49\linewidth}
\centering
\includegraphics[width=\linewidth]{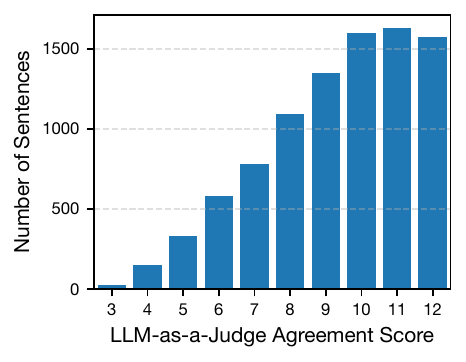}
\caption{}
\label{fig:judge_agreement_distribution}
\end{subfigure}
\hfill
\begin{subfigure}[t]{0.49\linewidth}
\centering
\includegraphics[width=\linewidth]{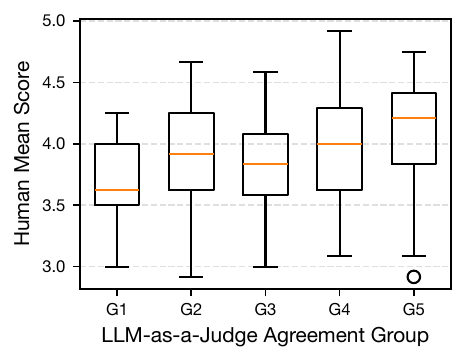}
\caption{}
\label{fig:human_groups_validation}
\end{subfigure}
\caption{
Sentence quality evaluation based on LLM-as-a-Judge agreement.
(a) Distribution of agreement scores over 9,108 sentences.
(b) Distribution of sentence-level human evaluation scores in each agreement group.
}
\Description{
Two plots summarizing sentence quality validation. Panel (a) shows the distribution of LLM-as-a-Judge agreement scores. Panel (b) compares human evaluation scores across the five agreement groups.
}
\end{figure}

Figure~\ref{fig:judge_agreement_distribution} shows the distribution of LLM-as-a-Judge agreement scores. The scores range from 3 to 12; the lowest scores are grouped as G1 (3--6) to ensure sufficient sample size. For the sentence-level human validation shown in Figure~\ref{fig:human_groups_validation}, we draw a stratified sample of 100 sentences: 10 from G1, 15 from G2, 25 from G3, 20 from G4, and 30 from G5. Each sentence-level score is computed by averaging the three annotators' five-point ratings over the four criteria.

\subsection{Auxiliary Metric Results}
\label{sec:app_intensity_stats}

Table~\ref{tab:intensity_stats} reports per-model means and standard deviations for the auxiliary YOLO metrics computed on images in which the metaphorical vehicle is detected.

\begin{table}[t]
\caption{
Bounding-box area ratio and detection confidence for detected metaphorical vehicles by model. 
}
\label{tab:intensity_stats}
\centering
\footnotesize
\renewcommand{\arraystretch}{1.15}
\setlength{\tabcolsep}{13.5pt}
\rowcolors{2}{gray!10}{white}

\makebox[\linewidth][c]{%
\begin{tabular}{lcccc}
\toprule
\multirow{2}{*}{\textbf{Model}} &
\multicolumn{2}{c}{\textbf{Area Ratio}} &
\multicolumn{2}{c}{\textbf{Confidence}} \\
\cmidrule(lr){2-3} \cmidrule(lr){4-5}
& \textbf{Mean} & \textbf{SD} & \textbf{Mean} & \textbf{SD} \\
\midrule
Dreamlike  & 0.1397 & 0.1838 & 0.7367 & 0.2044 \\
FLUX       & 0.1155 & 0.1427 & 0.8191 & 0.1795 \\
PixArt     & 0.1305 & 0.1513 & 0.7591 & 0.1932 \\
Qwen-Image & 0.1471 & 0.1687 & 0.7866 & 0.1807 \\
SD3.5      & 0.1335 & 0.1616 & 0.7746 & 0.1969 \\
\bottomrule
\end{tabular}
}
\end{table}

\section{Human Evaluation Instructions for Generated Images}
\label{sec:app_instruction}

The instructions shown to annotators are as follows.

\begin{promptbox}
\footnotesize\ttfamily
In this task, you will evaluate images generated by different models based on a given sentence containing a metaphor. For each sentence, you will see five images labeled A--E. The labels do not correspond to model names. Please evaluate each image independently according to the questions provided. There are no right or wrong answers. Focus on how well each image reflects the sentence and the metaphorical vehicle, rather than overall image quality or personal preference. The task consists of multiple sections and may take some time. Please work at a comfortable pace.
\end{promptbox}

\section{Diffusion Lens Visualization Examples}
\label{sec:app_difflens}

Table~\ref{tab:app_difflens_examples} presents representative Diffusion Lens visualizations. The examples cover all text encoder branches analyzed in the main paper and all seven base template types before expansion into singular and plural variants.

\begin{table*}[p]
\caption{
Representative Diffusion Lens visualization examples. Rows show one model or text encoder branch with its prompt, three layer-wise images, and the final image. Parentheses give layer/total layers; green boxes show YOLO detections with category/confidence labels; bold text marks simile expressions. 
}
\label{tab:app_difflens_examples}
\centering
\scriptsize
\renewcommand{\arraystretch}{0.95}
\resizebox{\textwidth}{!}{%
\begin{tabular}{>{\centering\arraybackslash}m{0.15\textwidth}|>{\centering\arraybackslash}m{0.21\textwidth}>{\centering\arraybackslash}m{0.21\textwidth}>{\centering\arraybackslash}m{0.21\textwidth}>{\centering\arraybackslash}m{0.21\textwidth}}
\toprule
\textbf{Model} & \multicolumn{4}{c}{\textbf{Layer-Wise Visualization Images and Final Image}} \\
\midrule

\multicolumn{5}{l}{\textbf{Examples of appearance in shallow layers followed by persistence to the final image}} \\
\midrule
\textbf{} & \multicolumn{4}{c}{\small\textit{The old man sat, \textbf{as} still \textbf{as a potted plant}, on the park bench.}} \\[2pt]
\textbf{FLUX (CLIP)} &
\shortstack{\includegraphics[width=1.75cm]{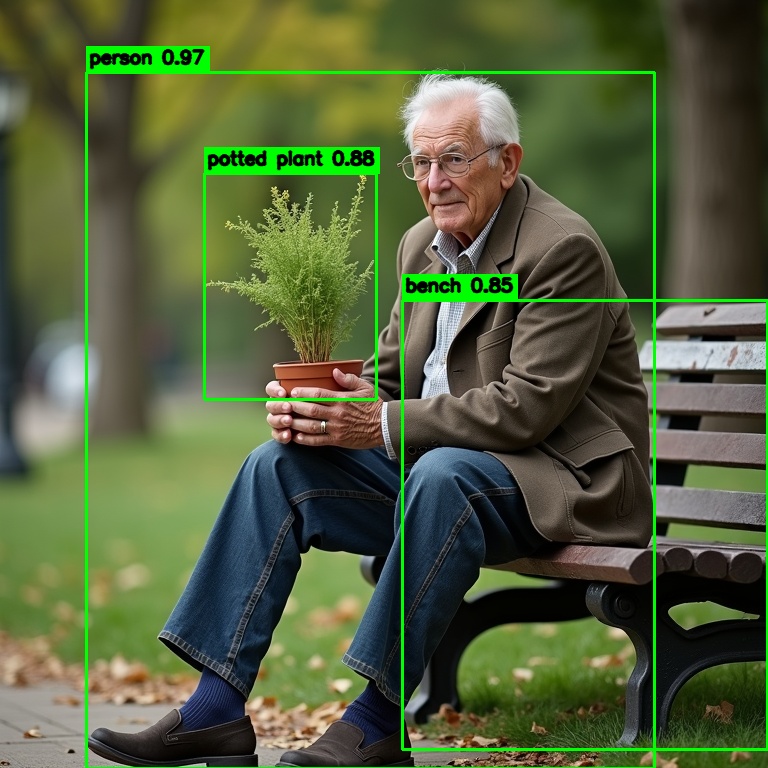}\\(2/12)} &
\shortstack{\includegraphics[width=1.75cm]{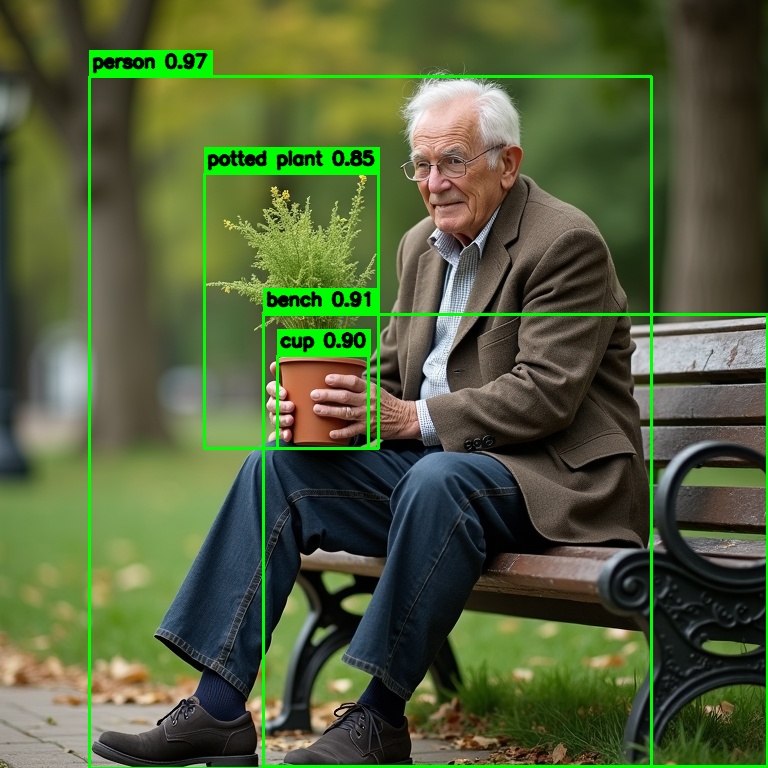}\\(6/12)} &
\shortstack{\includegraphics[width=1.75cm]{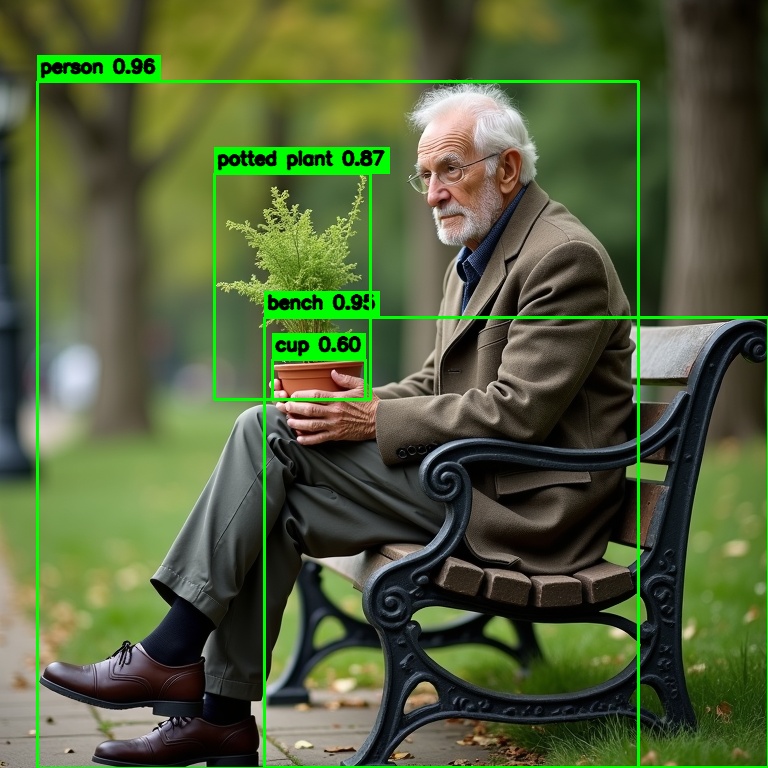}\\(12/12)} &
\shortstack{\includegraphics[width=1.75cm]{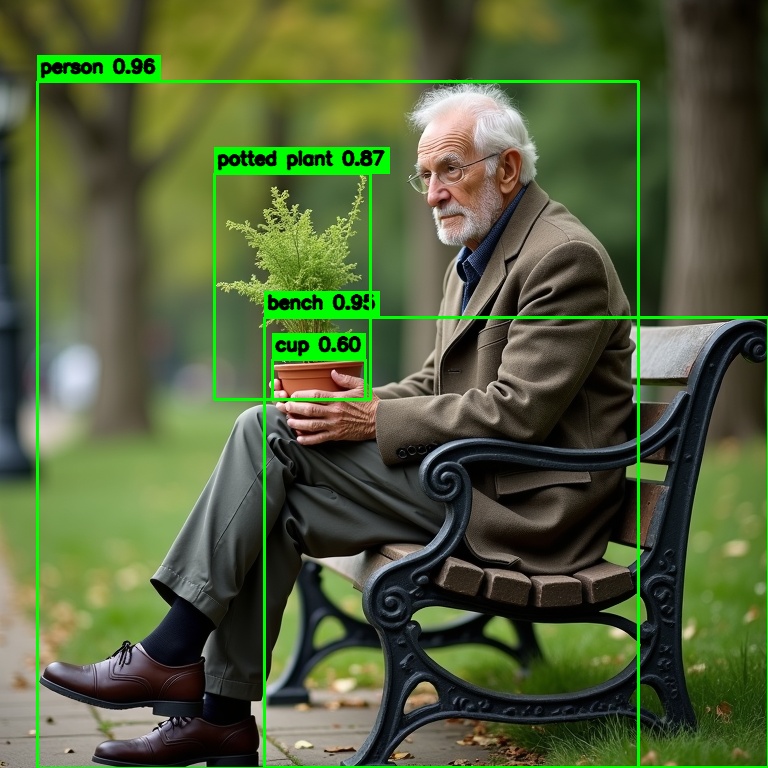}\\(Final)} \\
\midrule
\textbf{} & \multicolumn{4}{c}{\small\textit{The clouds drifted \textbf{exactly like horses} across the sky.}} \\[2pt]
\textbf{FLUX (T5)} &
\shortstack{\includegraphics[width=1.75cm]{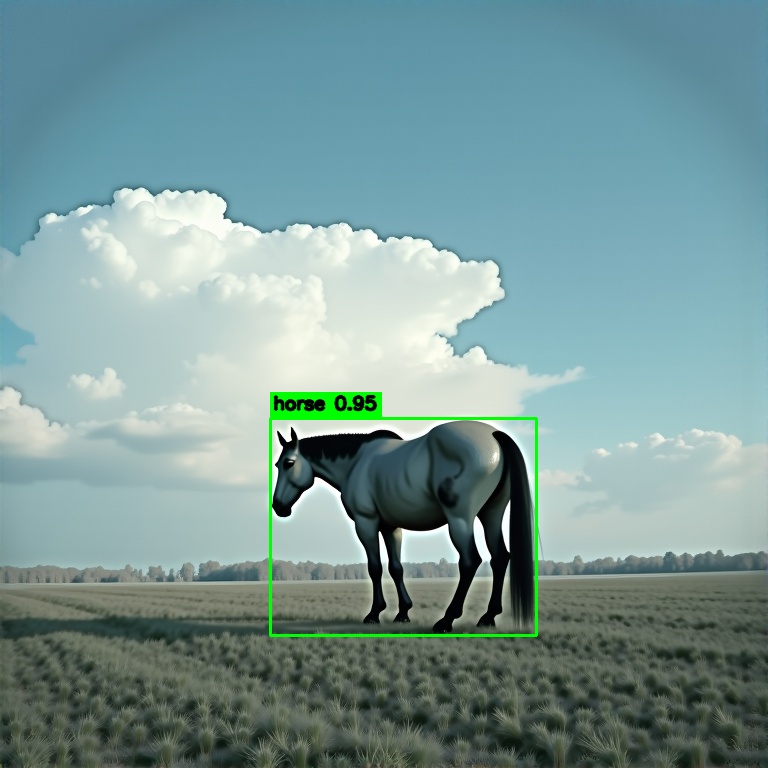}\\(1/24)} &
\shortstack{\includegraphics[width=1.75cm]{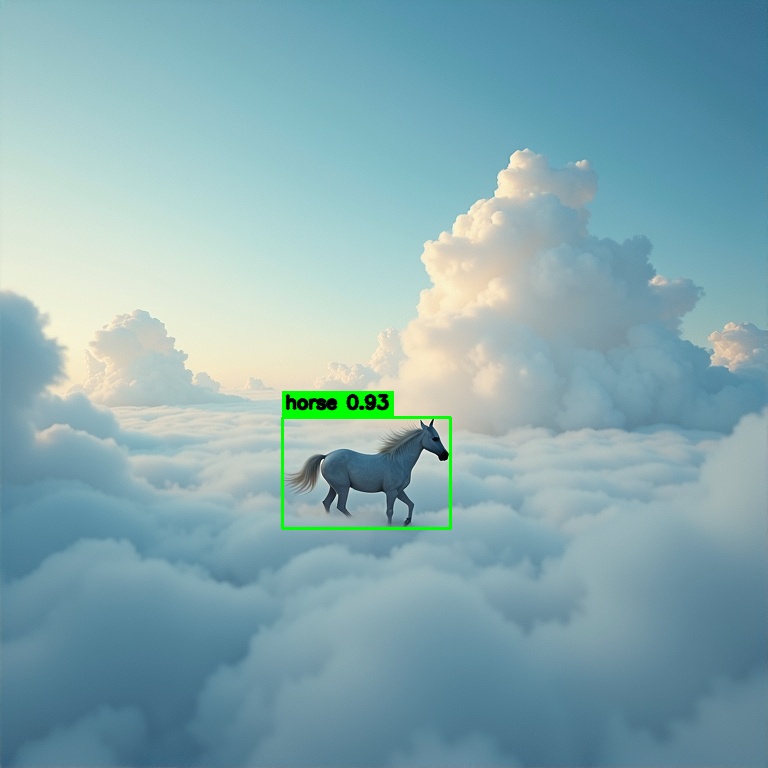}\\(12/24)} &
\shortstack{\includegraphics[width=1.75cm]{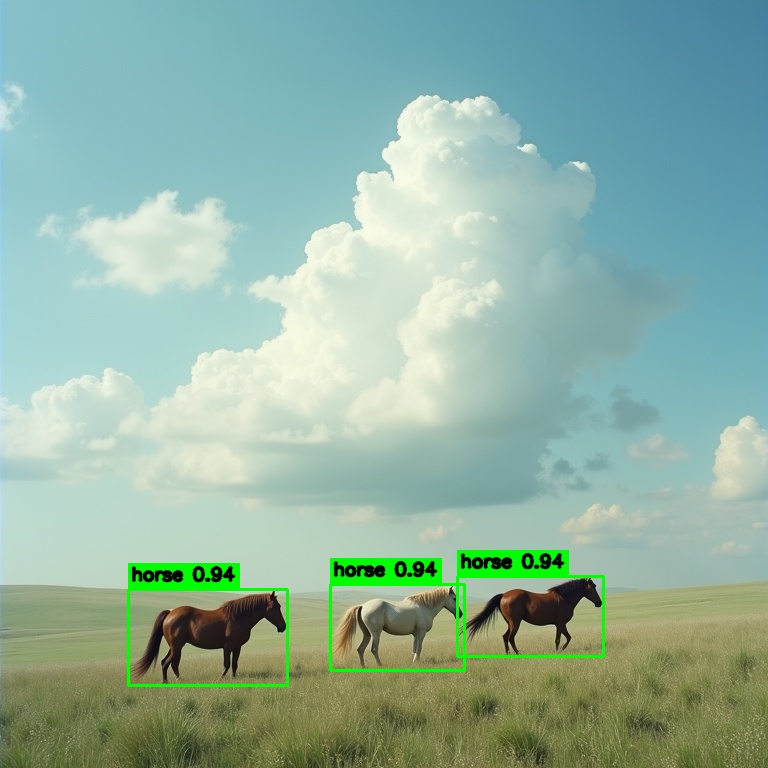}\\(24/24)} &
\shortstack{\includegraphics[width=1.75cm]{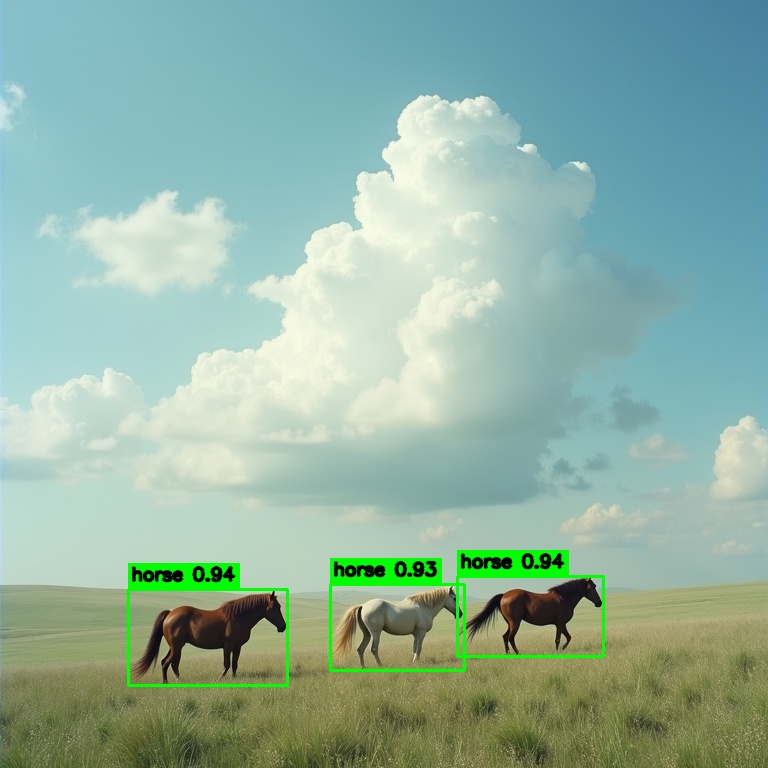}\\(Final)} \\
\midrule
\textbf{} & \multicolumn{4}{c}{\small\textit{The rocks \textbf{look like benches} scattered along the riverbank.}} \\[2pt]
\textbf{SD3.5 (CLIP-L)} &
\shortstack{\includegraphics[width=1.75cm]{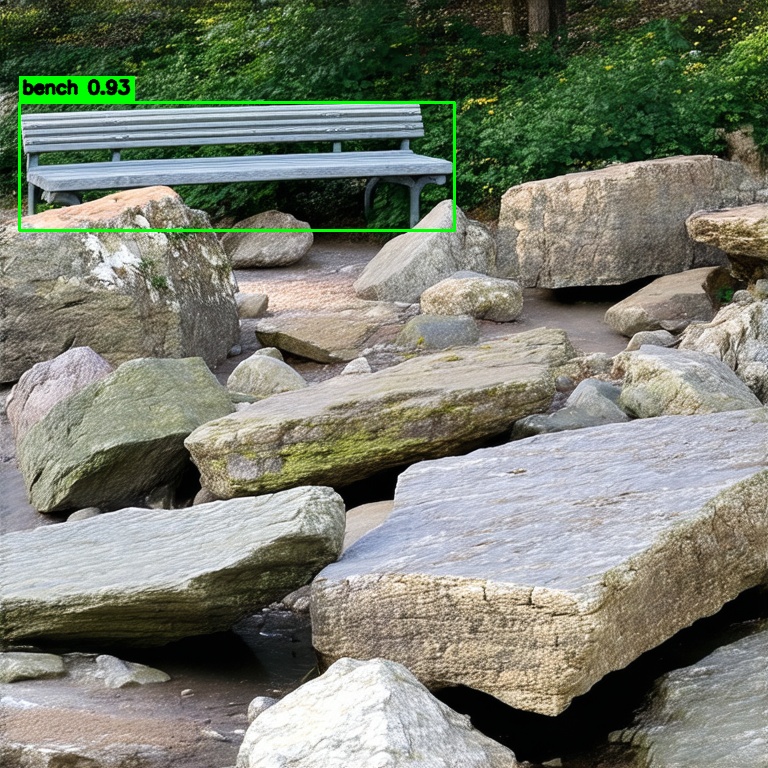}\\(1/12)} &
\shortstack{\includegraphics[width=1.75cm]{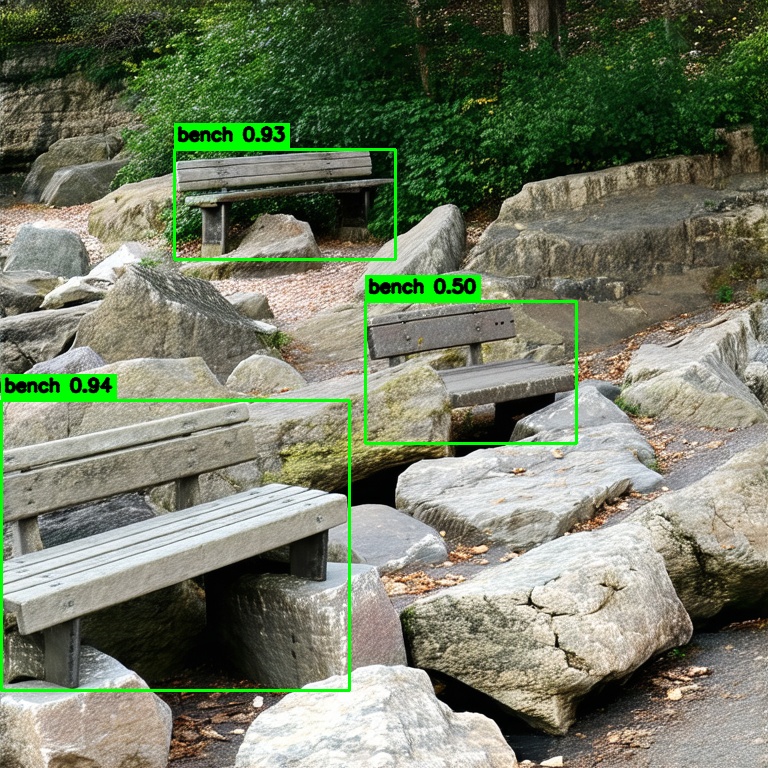}\\(6/12)} &
\shortstack{\includegraphics[width=1.75cm]{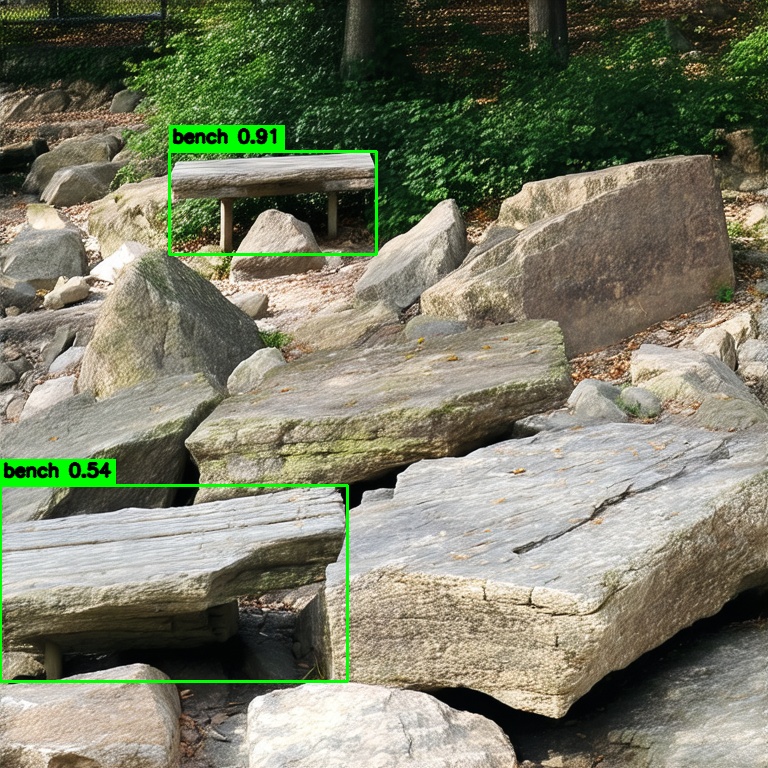}\\(12/12)} &
\shortstack{\includegraphics[width=1.75cm]{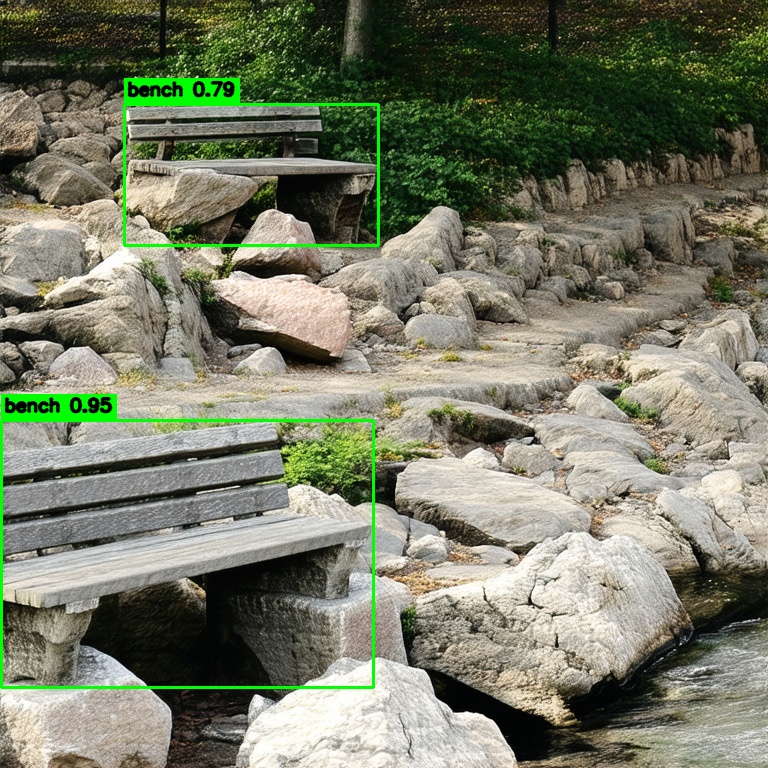}\\(Final)} \\

\midrule
\multicolumn{5}{l}{\textbf{Examples of appearance in intermediate layers followed by suppression before the final image}} \\
\midrule
\textbf{} & \multicolumn{4}{c}{\small\textit{The wind cut through the trees \textbf{like knives}.}} \\[2pt]
\textbf{Dreamlike} &
\shortstack{\includegraphics[width=1.75cm]{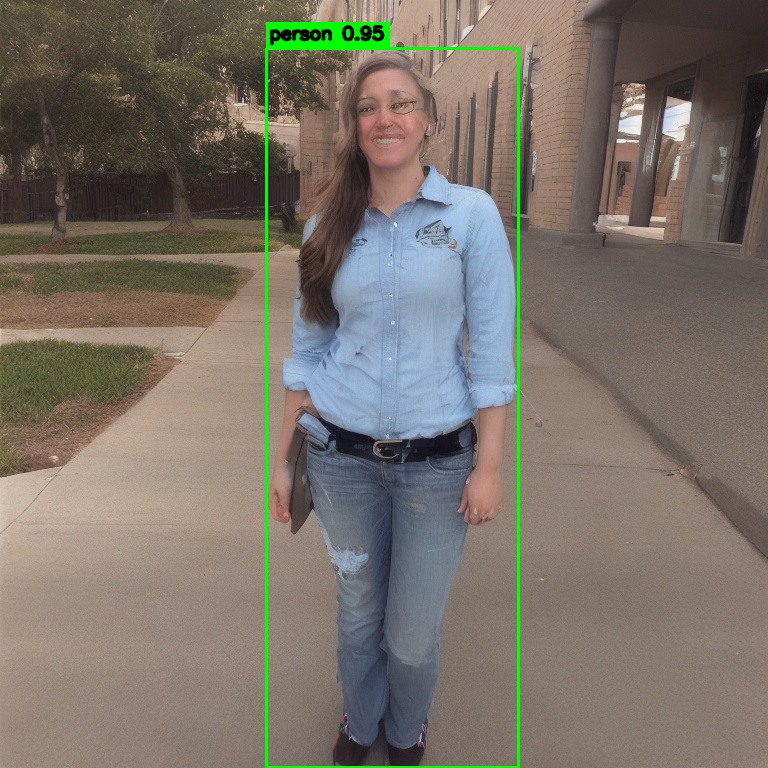}\\(1/12)} &
\shortstack{\includegraphics[width=1.75cm]{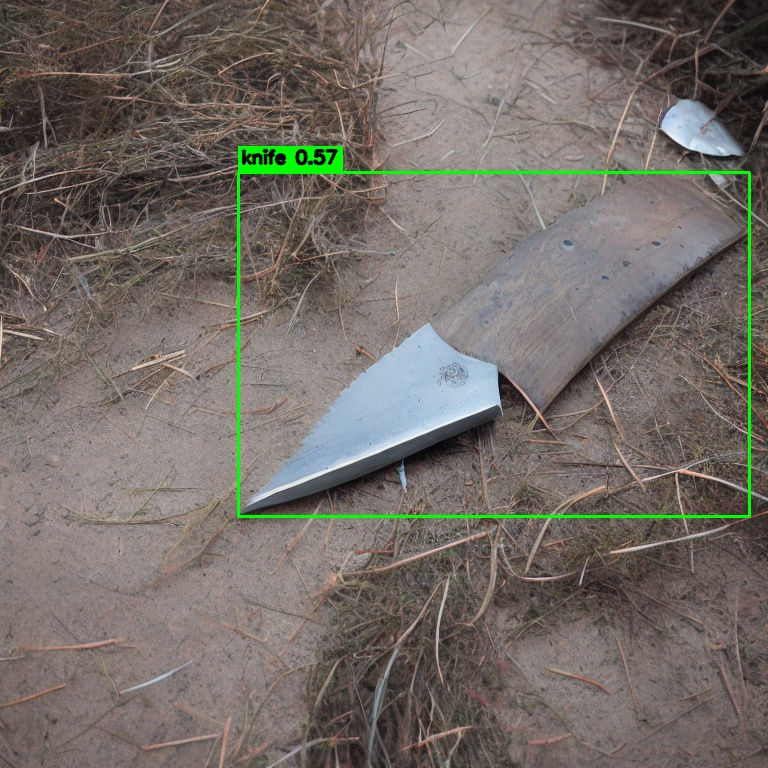}\\(6/12)} &
\shortstack{\includegraphics[width=1.75cm]{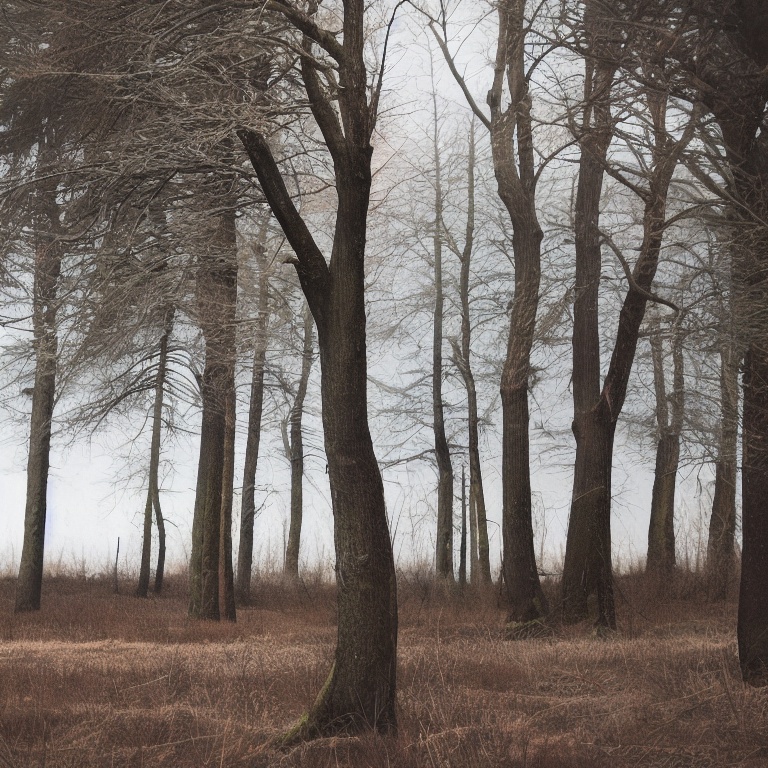}\\(12/12)} &
\shortstack{\includegraphics[width=1.75cm]{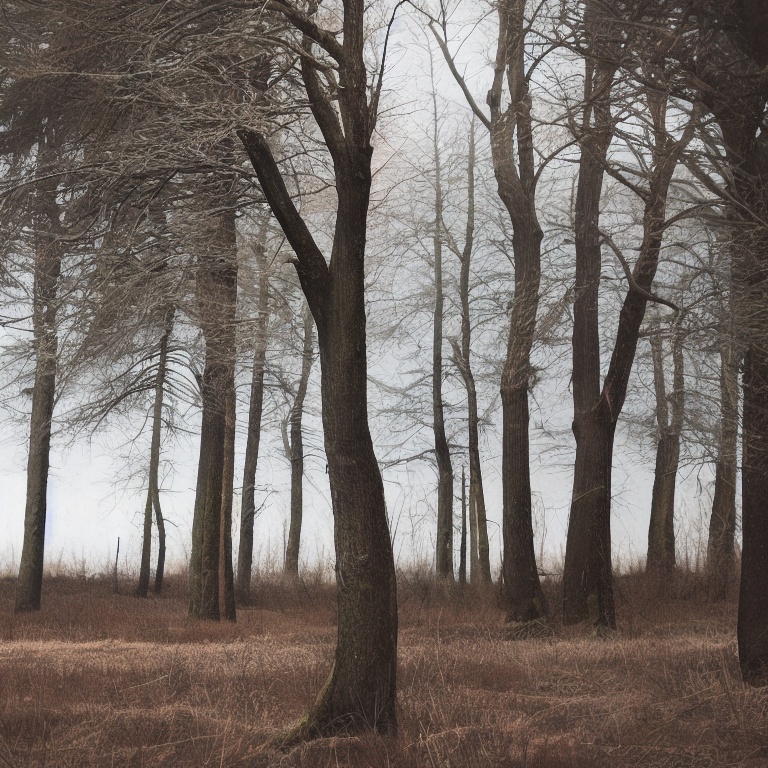}\\(Final)} \\
\midrule
\textbf{} & \multicolumn{4}{c}{\small\textit{The tree branch \textbf{looks like a baseball bat}.}} \\[2pt]
\textbf{SD3.5 (CLIP-G)} &
\shortstack{\includegraphics[width=1.75cm]{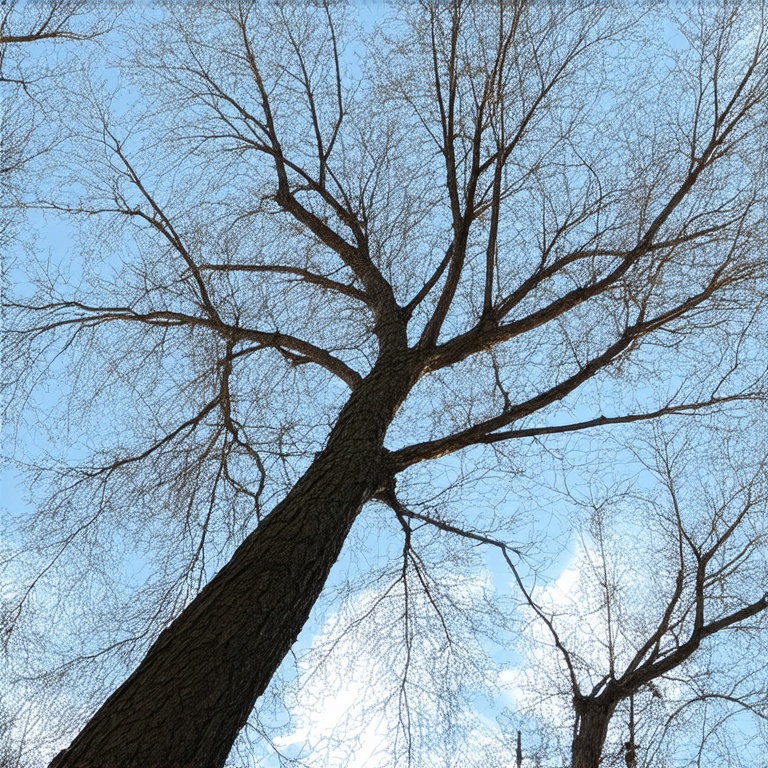}\\(1/32)} &
\shortstack{\includegraphics[width=1.75cm]{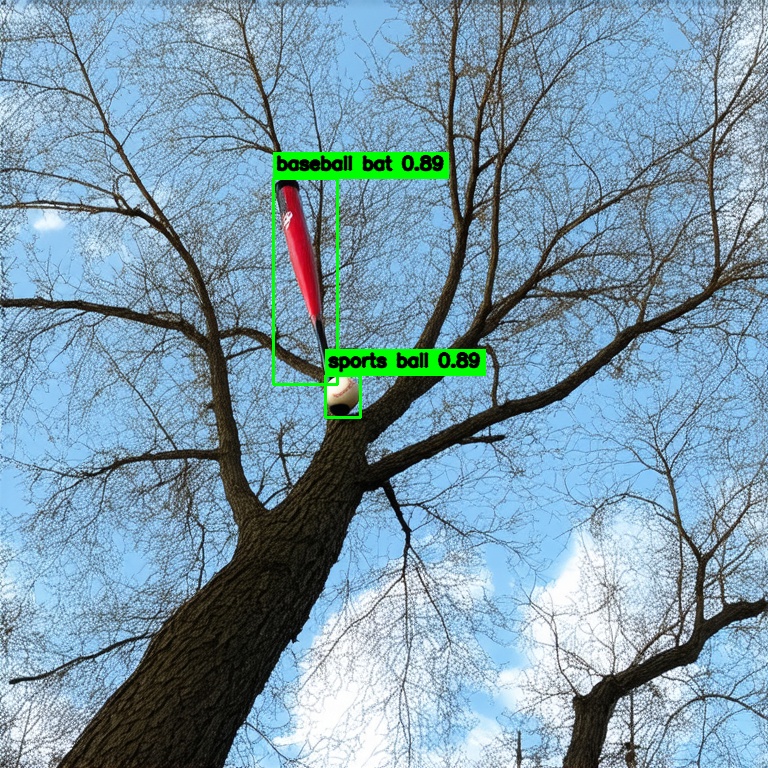}\\(5/32)} &
\shortstack{\includegraphics[width=1.75cm]{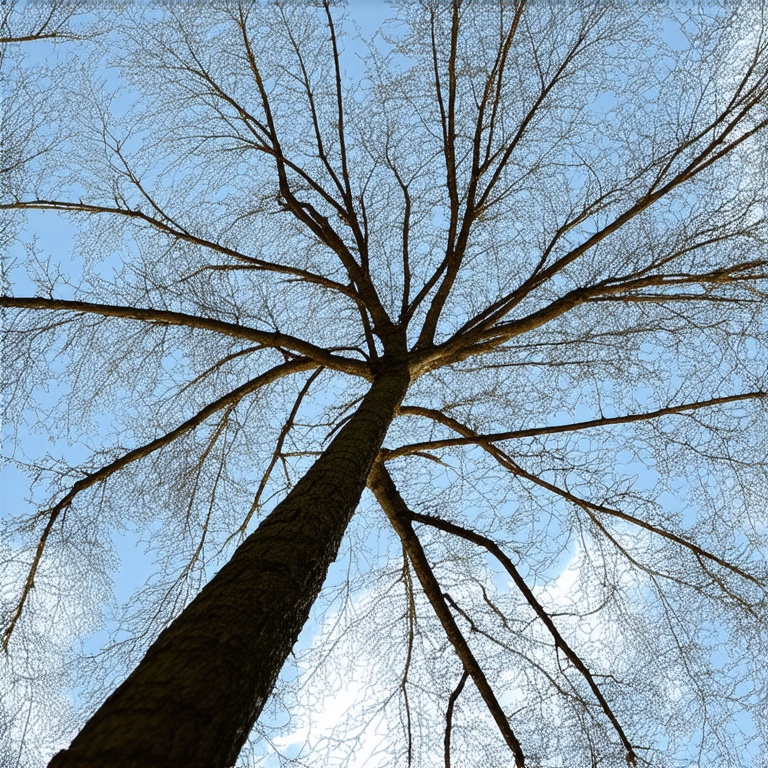}\\(32/32)} &
\shortstack{\includegraphics[width=1.75cm]{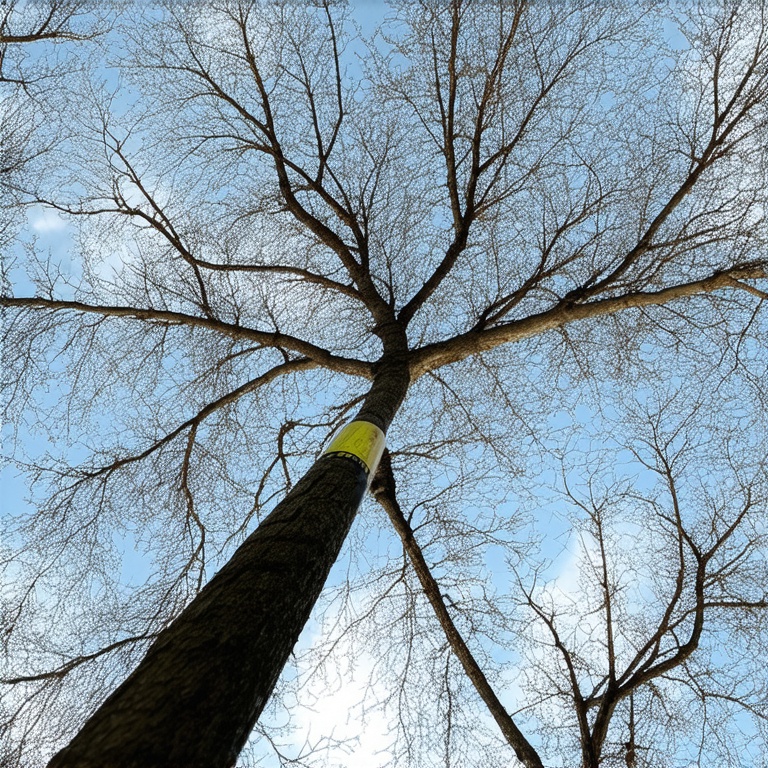}\\(Final)} \\
\midrule
\textbf{} & \multicolumn{4}{c}{\small\textit{The city lights twinkled below, \textbf{as though bowls} of scattered jewels had been spilled across the landscape.}} \\[2pt]
\textbf{SD3.5 (T5)} &
\shortstack{\includegraphics[width=1.75cm]{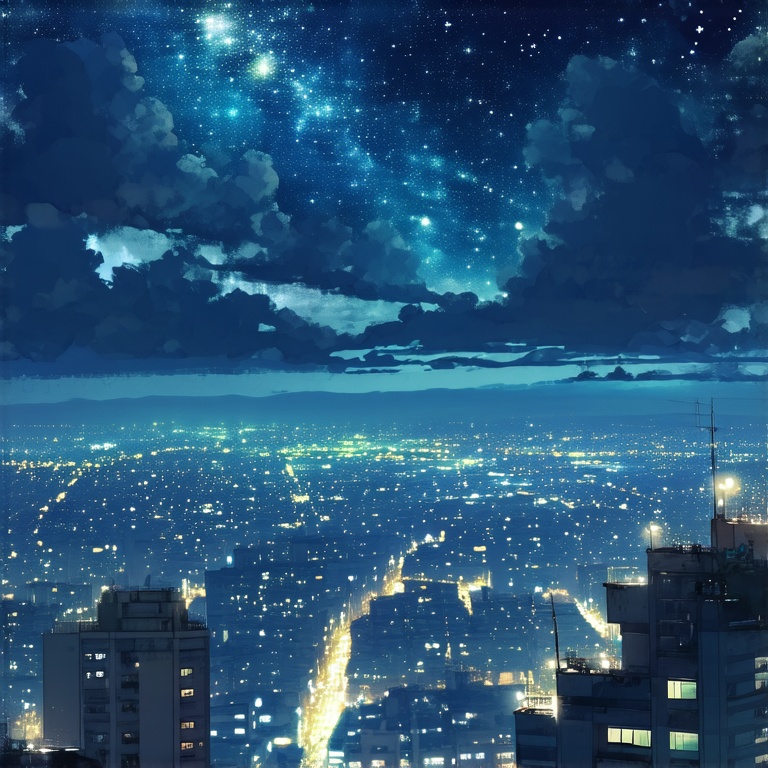}\\(1/24)} &
\shortstack{\includegraphics[width=1.75cm]{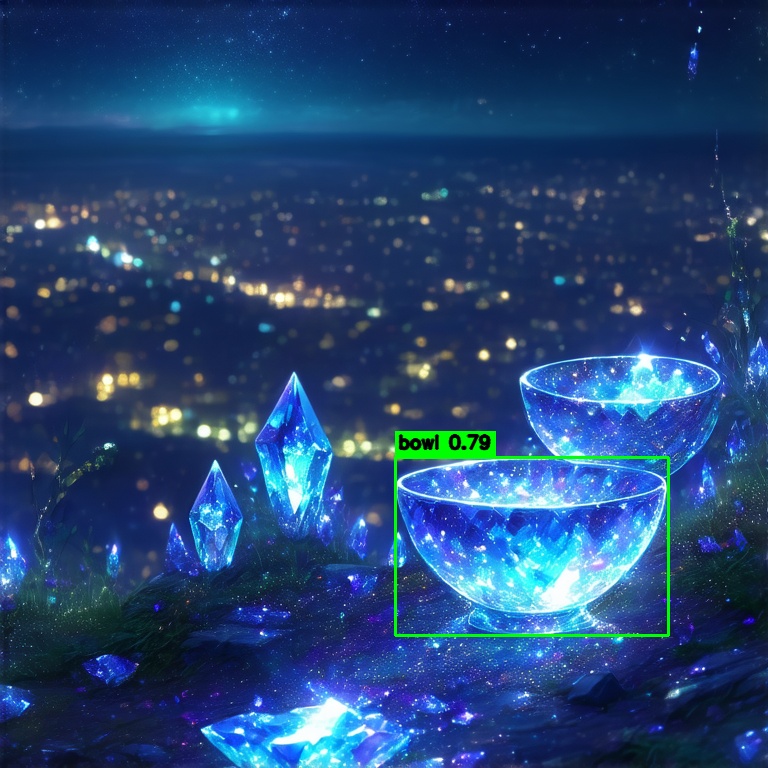}\\(17/24)} &
\shortstack{\includegraphics[width=1.75cm]{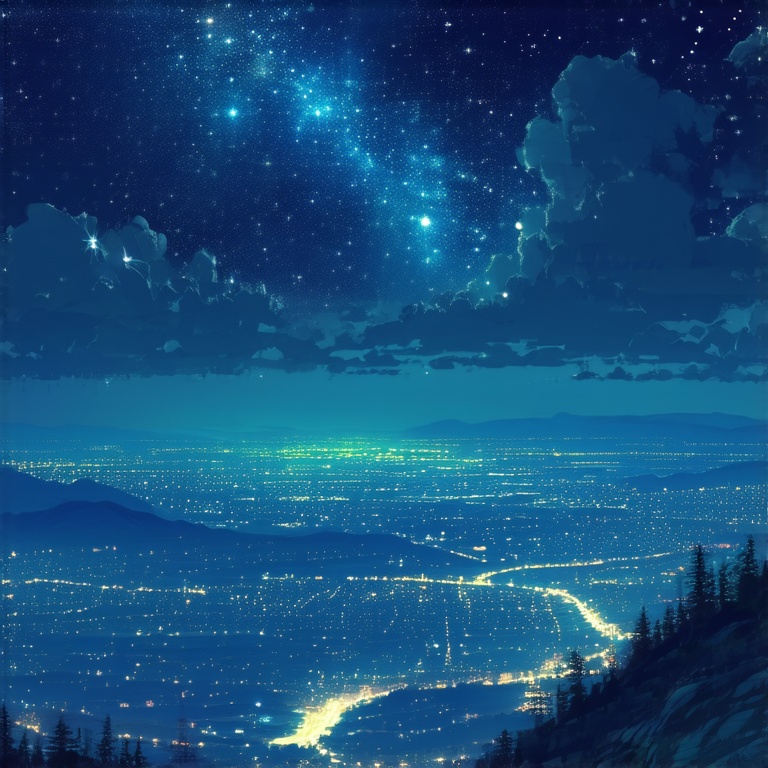}\\(24/24)} &
\shortstack{\includegraphics[width=1.75cm]{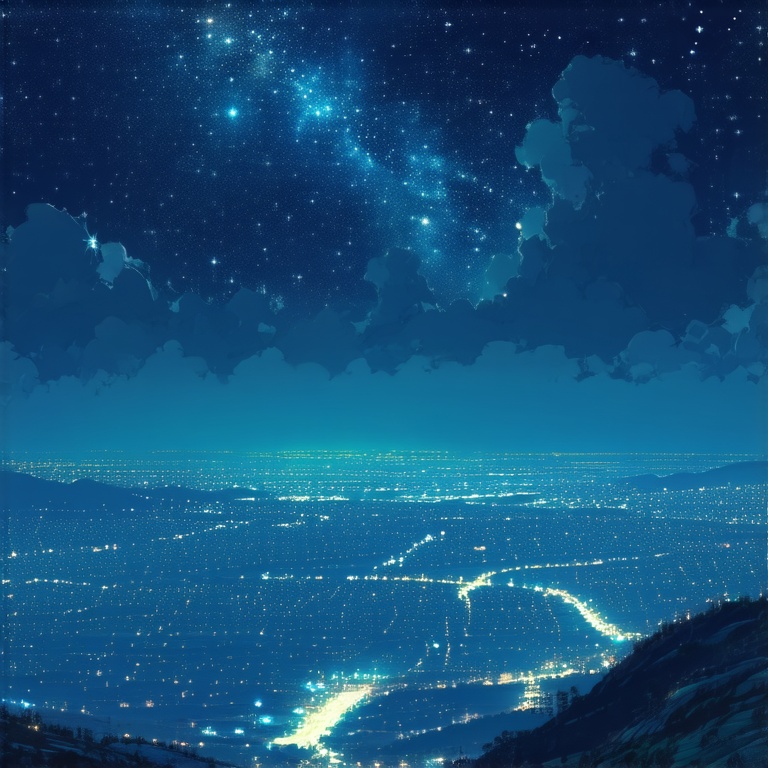}\\(Final)} \\

\midrule
\multicolumn{5}{l}{\textbf{Examples of appearance in deeper layers followed by persistence to the final image}} \\
\midrule
\textbf{} & \multicolumn{4}{c}{\small\textit{The tall buildings in the city skyline stretched \textbf{just like giraffes}, their necks reaching for the clouds.}} \\[2pt]
\textbf{PixArt} &
\shortstack{\includegraphics[width=1.75cm]{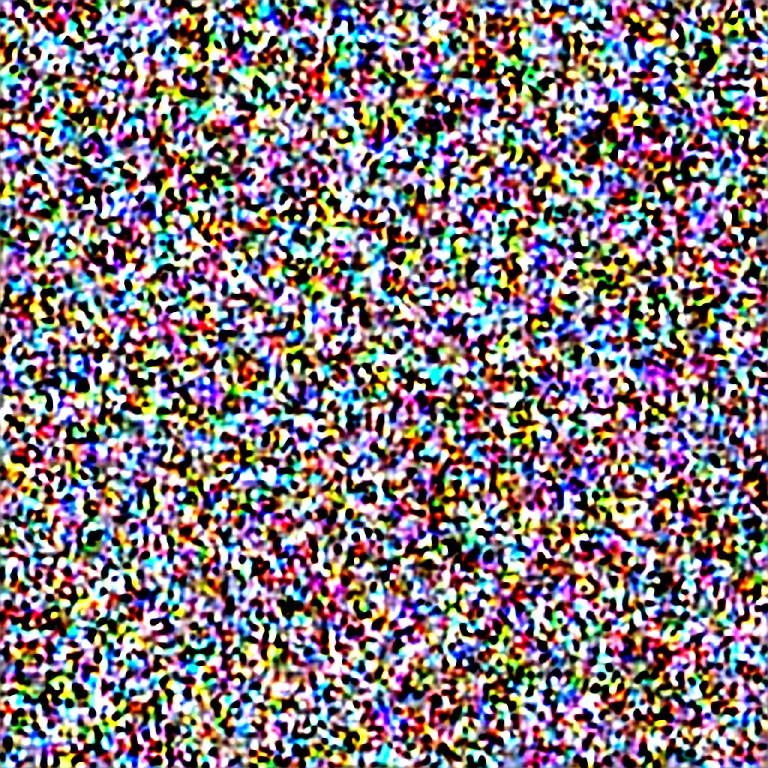}\\(1/24)} &
\shortstack{\includegraphics[width=1.75cm]{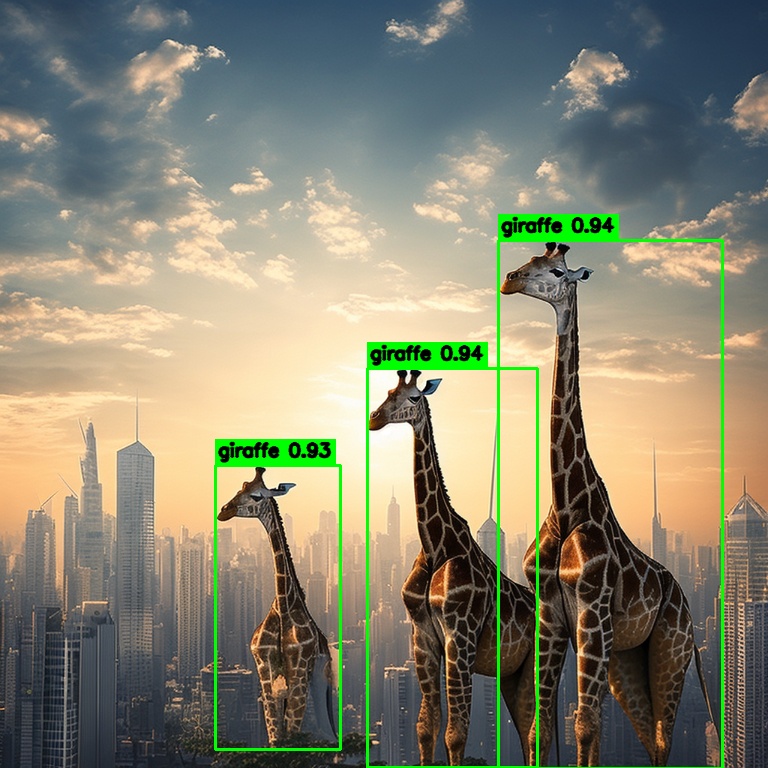}\\(22/24)} &
\shortstack{\includegraphics[width=1.75cm]{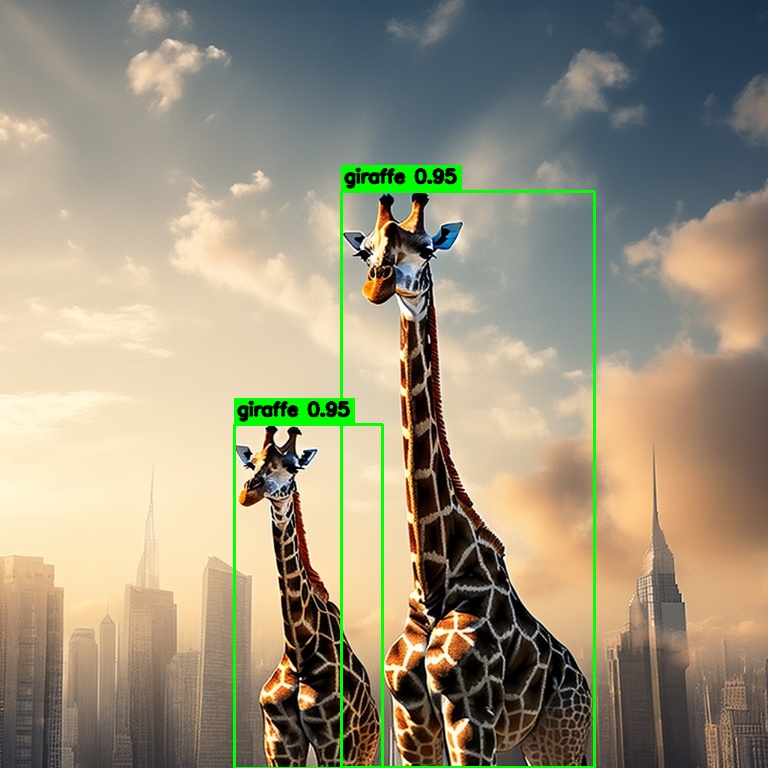}\\(24/24)} &
\shortstack{\includegraphics[width=1.75cm]{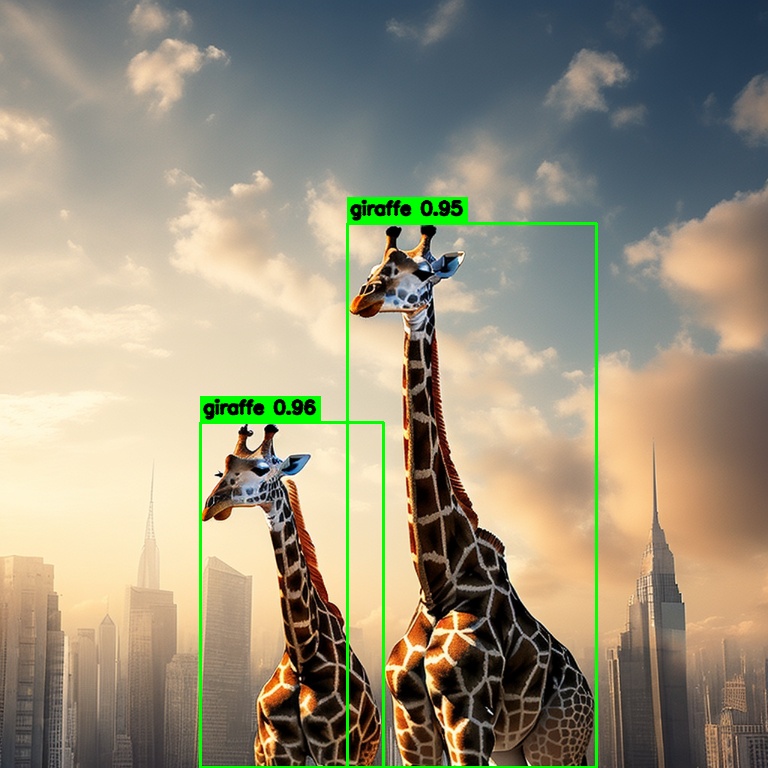}\\(Final)} \\
\midrule
\textbf{} & \multicolumn{4}{c}{\small\textit{The city streets were \textbf{as if dining tables} scattered with crumbs and spills, reflecting the chaos of the festival.}} \\[2pt]
\textbf{Qwen-Image} &
\shortstack{\includegraphics[width=1.75cm]{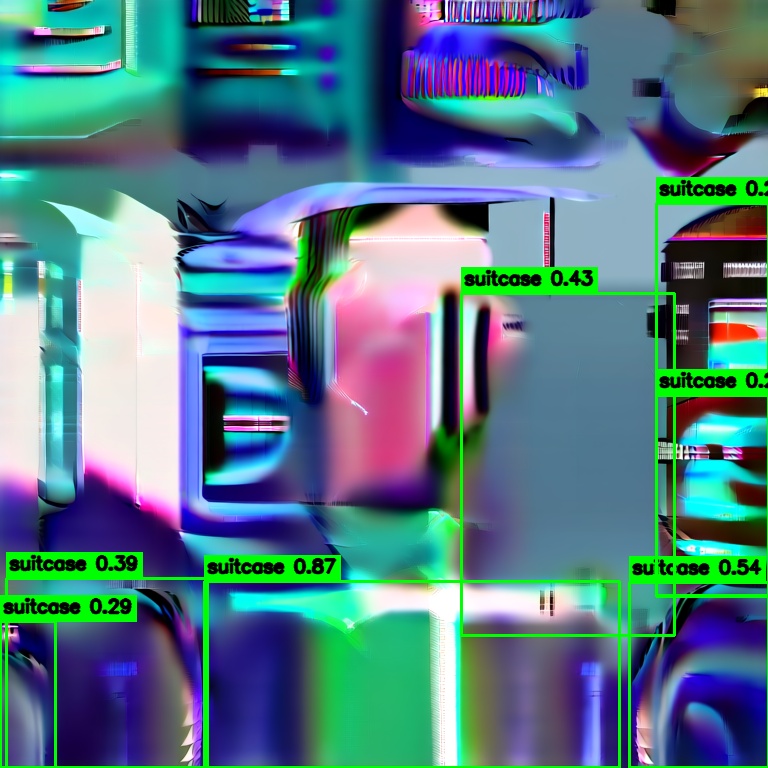}\\(1/28)} &
\shortstack{\includegraphics[width=1.75cm]{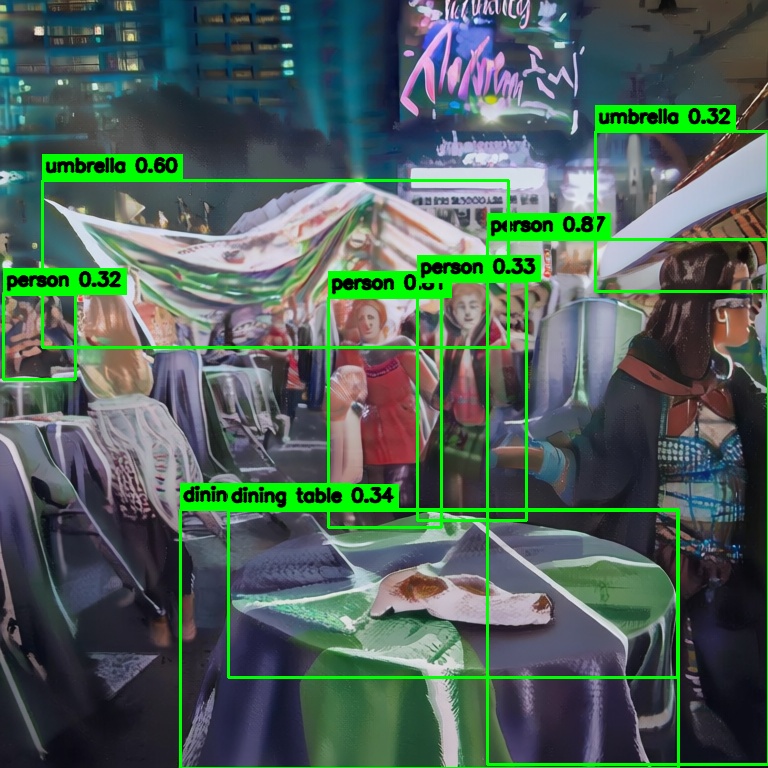}\\(23/28)} &
\shortstack{\includegraphics[width=1.75cm]{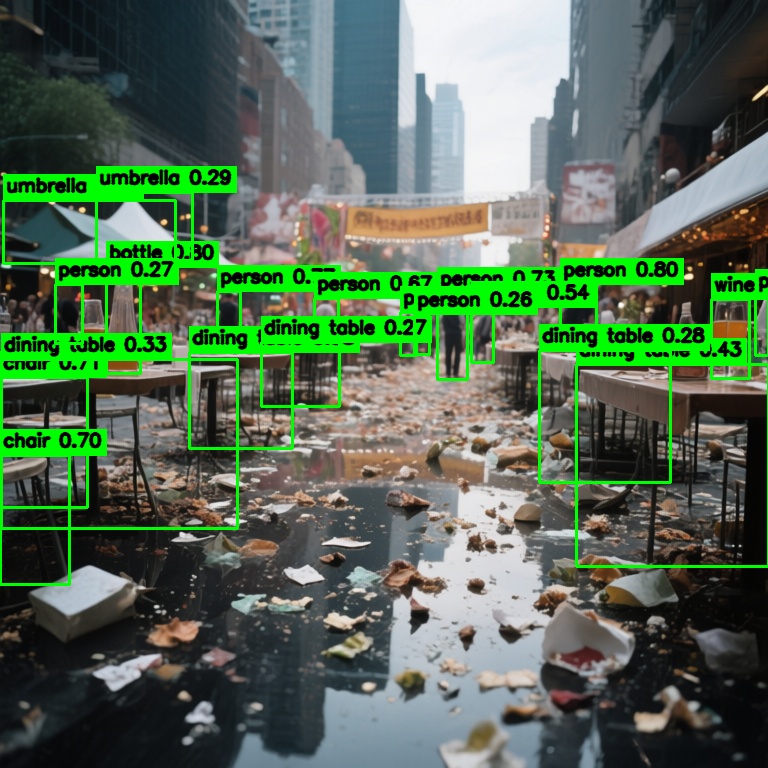}\\(28/28)} &
\shortstack{\includegraphics[width=1.75cm]{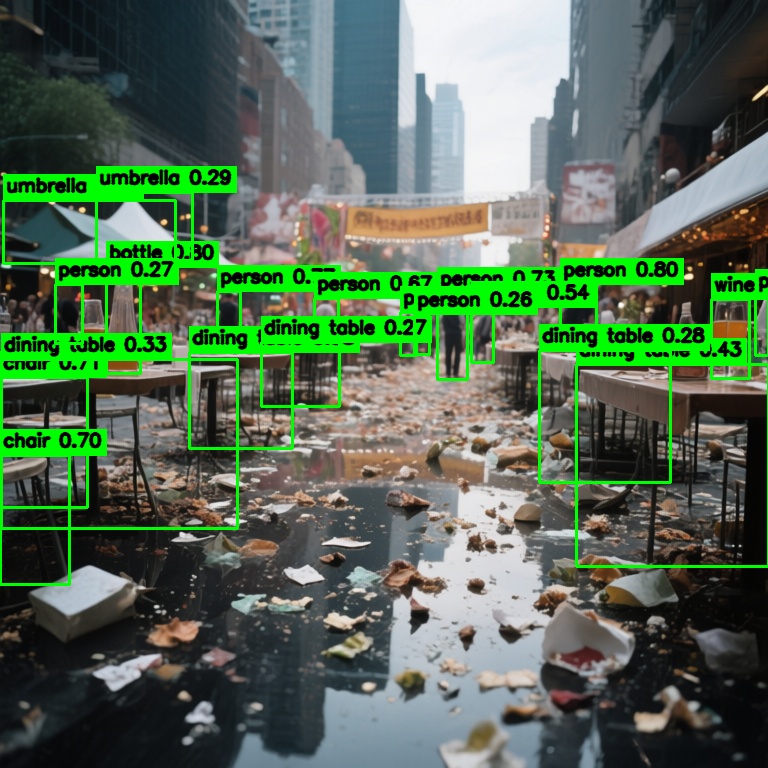}\\(Final)} \\

\bottomrule
\end{tabular}%
}
\end{table*}

\end{document}